\documentclass[conference]{IEEEtran}
\usepackage{times}

\usepackage[numbers]{natbib}
\usepackage{multicol}
\usepackage[bookmarks=true]{hyperref}

\IEEEoverridecommandlockouts

\usepackage{mathtools}
\usepackage{amssymb}
\usepackage{pifont}
\usepackage{hyperref}
\usepackage{makecell}
\usepackage[ruled, english, englishkw]{algorithm2e}
\usepackage{xspace}
\usepackage{multirow}
\usepackage{mathtools}
\usepackage{float}
\usepackage{array} 
\usepackage{longtable}
\usepackage{dcolumn}
\usepackage{threeparttable}
\usepackage{listings}
\usepackage{calc}
\usepackage{tabularx}
\usepackage{comment}
\usepackage{xtab}
\usepackage[table]{xcolor}
\usepackage{colortbl}
\usepackage{tikz}
\usetikzlibrary{arrows}
\usetikzlibrary{automata}

\makeatletter
\let\mcnewpage\newpage
\newcommand{\changenewpage}{%
  \renewcommand\newpage{%
    \if@firstcolumn
      \hrule width\linewidth height0pt
      \columnbreak
    \else
      \mcnewpage
    \fi
}}
\makeatother

\usepackage[inline]{enumitem} % for inline enumerates 

\usepackage{balance}
\usepackage{booktabs}

\usepackage{graphicx}
\usepackage{subcaption}

\usepackage[inline]{enumitem}
\usepackage{soul} 

\usepackage[per-mode=fraction]{siunitx}
\usepackage{comment}
\usepackage[normalem]{ulem}
\usepackage{xspace}

\usepackage{cancel}
\usepackage{hyperref}

\DeclareSIUnit\px{px}
\DeclareSIUnit\fps{fps}

\definecolor{OliveGreen}{RGB}{0,200,25}
\newcommand{\red}[1]{\textcolor{red}{#1}}

\newcommand{\darkgreen}[1]{\textcolor{OliveGreen}{#1}}
\newcommand{\blue}[1]{\textcolor{blue}{#1}}
\newcommand{\orange}[1]{\textcolor{orange}{#1}}

\newcommand{\ie}{i.\,e.,\xspace}
\newcommand{\eg}{e.\,g.,\xspace}

\newcommand{\armarVI}{\mbox{ARMAR-6}\xspace}

\newcommand{\armarDE}{\mbox{ARMAR-DE}\xspace}

\newcommand{\ackHariaEuRobinJuBot}{The research leading to these results has received funding from the European Union’s Horizon Europe programme under grant agreement No. 101070292 (HARIA) and No. 101070596 (euROBIN) and from the Carl Zeiss Foundation through the JuBot project.}

\newcommand{\added}[1]{\darkgreen{#1}}
\newcommand{\replaced}[2]{\red{\ifmmode\text{\sout{\ensuremath{#1}}}\else\sout{#1}\fi}\;\darkgreen{#2}}
\newcommand{\removed}[1]{\red{\ifmmode\text{\sout{\ensuremath{#1}}}\else\sout{#1}\fi}}

\newcommand{\remark}[1]{\blue{--- #1 ---}}
\newcommand{\todo}[1]{\{\orange{---TODO--- #1}\}}
\newcommand{\toremove}[1]{#1}

\newif\iffinal
\iffinal
	\renewcommand{\added}[1]{#1}
	\renewcommand{\replaced}[2]{#2}
	\renewcommand{\removed}[1]{}
	\renewcommand{\remark}[1]{}
	\renewcommand{\todo}[1]{}
	\renewcommand{\toremove}[1]{}
\fi

\newcommand{\removedfootnote}[1]{\footnote{\removed{#1}}}
\newcommand{\removedsubsection}[1]{\subsection{\texorpdfstring{\removed{#1}}{#1}}}

\newcommand{\addedsubsection}[1]{\subsection{\texorpdfstring{\added{#1}}{#1}}}

\iffinal
	
	\renewcommand{\removedfootnote}[1]{}
	\renewcommand{\removedsubsection}[1]{}
	
	\renewcommand{\removedsubsection}[1]{}

\fi

\newcolumntype{M}[1]{>{\centering\arraybackslash}m{#1}}

\newif\iffinal
\finaltrue

\iffinal
	\renewcommand{\added}[1]{#1} 
	\renewcommand{\replaced}[2]{#2}
	\renewcommand{\removed}[1]{}
	\renewcommand{\remark}[1]{}
\fi

\newif\ifdoubleblind

\newcommand{\ChatGPT}{\emph{ChatGPT}\xspace}
\newcommand{\AutoGPTP}{\emph{AutoGPT+P}\xspace}
\newcommand{\SayCan}{\emph{SayCan}\xspace}
\newcommand{\PDDL}{PDDL\xspace}
\newcommand{\LLMP}{\emph{LLM+P}\xspace}
\newcommand{\LLMasPlanner}{\emph{LLM as Planner}\xspace}
\newcommand{\LLMwPlanner}{\emph{LLM with Planner}\xspace}
\newcommand{\OAM}{\emph{Object Affordance Mapping}\xspace}
\newcommand{\pickplace}{picking and placing\xspace}
\newcommand{\handover}{handover\xspace}
\newcommand{\pour}{pouring\xspace}
\newcommand{\chop}{chopping\xspace}
\newcommand{\heat}{heating\xspace}
\newcommand{\wipe}{wiping\xspace}
\newcommand{\GPTIII}{\emph{GPT-3}\xspace}
\newcommand{\GPTIV}{\emph{GPT-4}\xspace}

\ifdoubleblind
    \renewcommand{\armarVI}{the actual robot}
\fi

\begin{document}

\title{AutoGPT+P: Affordance-based Task Planning \\using Large Language Models
}

% You will get a Paper-ID when submitting a pdf file to the conference system
\author{Timo Birr, Christoph Pohl, Abdelrahman Younes and Tamim Asfour\\
%\author{\authorblockN{Michael Shell}
\small{Karlsruhe Institute of Technology, Germany}\\
\small{ \{timo.birr, asfour\}@kit.edu}}

% avoiding spaces at the end of the author lines is not a problem with
% conference papers because we don't use \thanks or \IEEEmembership

%Homer Simpson\authorrefmark{2},
%Montgomery Scott\authorrefmark{3} and
%Eldon Tyrell\authorrefmark{4}}
%Telephone: (800) 555--1212, Fax: (888) 555--1212}
%\authorblockA{\authorrefmark{4}Tyrell Inc., 123 Replicant Street, Los Angeles, California 90210--4321}}

\maketitle

\begin{abstract}
%Recent works in planning try to leverage the advances in \emph{Large Language Models} (LLMs) by combining them with classical planners, thereby overcoming the limited reasoning capabilities of LLMs. 
Recent advances in task planning leverage Large Language Models (LLMs) to improve generalizability by combining such models with classical planning algorithms to address their inherent limitations in reasoning capabilities. %Currently, they are limited by not being integrated into a system to dynamically capture the initial state of the planning problem. 
However, these approaches face the challenge of dynamically capturing the initial state of the task planning problem. To alleviate this issue, we propose \AutoGPTP, a system that combines an affordance-based scene representation with a planning system. Affordances are the action possibilities of an agent on the environment and the objects present in it. Thus, deriving the planning domain from an affordance-based scene representation allows symbolic planning with arbitrary objects. \AutoGPTP leverages this representation to derive and execute a plan for a task specified by the user in natural language. In addition to solving planning tasks under a closed-world assumption, \AutoGPTP can also handle planning with incomplete information, such as tasks with missing objects, by exploring the scene, suggesting alternatives, or providing a partial plan. The affordance-based scene representation combines object detection with \replaced{an automatically generated \OAM using \ChatGPT.}{an \OAM that is  automatically generated using \ChatGPT.} The core planning tool extends existing work by automatically correcting semantic and syntactic errors\replaced{.Our approach achieves a success rate of 98\%, surpassing the current 81\% success rate of the current state-of-the-art LLM-based planning method \SayCan}{ leading to a success rate of 98\%} on the \SayCan instruction set. Furthermore, we evaluated our approach on our newly created dataset with 150 scenarios covering a wide range of complex tasks with missing objects, achieving a success rate of 79\%.\removed{on our dataset.} The dataset and the code are publicly available at \url{https://git.h2t.iar.kit.edu/sw/autogpt-p}.%\AutoGPTP shows the ability to generate plans for complicated tasks and achieves an average success rate of 79\% on our dataset containing 150 tasks, making it a versatile and flexible planning system.
\end{abstract}

\IEEEpeerreviewmaketitle
 \section{Introduction}
%For humanoid robots to improve as assistants, intuitive interaction in natural language is crucial.
The effectiveness of natural language interaction between humans and robots has been empirically confirmed as highly efficient \cite{liu2019review}. For example, Kartmann et al. \cite{Kartmann2021, Kartmann2023} demonstrate the incremental learning of spatial relationships through demonstrations and reproducing the learned relationships through natural language commands, providing an intuitive way to manipulate scenes semantically. Despite the recent notable advancements in Natural Language Processing (NLP) and understanding, particularly with the emergence of \emph{Large Language Models} (LLMs)\removed{, which have gained significant attention in cognitive computing research and proven effective as zero-shot learners}, these models still face limitations. Specifically, LLMs currently lack the ability to directly translate a natural language instruction into a plan for executing robotic tasks, primarily due to their constrained reasoning capabilities \cite{valmeekam2023large,MichaelDosaycan}.
\added{Recently, \LLMP \cite{liu2023llmp} demonstrated the capacity for enhancing the planning capabilities of LLMs by combining them with classical planners, grounding them with the planning domain and objects within the scene. However, the system is restricted by the closed-world assumption of classical planners. Thus, it can only generate plans if all objects needed to complete the task are available. Furthermore, \LLMP has no automated error correction and is vulnerable to contradictory goal definitions of the LLM.}

\replaced{In this work}{To overcome these restrictions}, we introduce \AutoGPTP, a system that enables users to command robots in natural language, derive and execute a plan to fulfill the user's request even if the objects needed to perform the task are missing in the immediate environment. \AutoGPTP exhibits dynamic responsiveness by searching the environment for missing objects, proposing alternatives, or progressing towards a subgoal when faced with such constraints.

\begin{figure}[!t]
 \usetikzlibrary{positioning,shapes}
    \centering
    \tikzset{subcaption/.style={fill=black,text=white, minimum width=0.31\linewidth},
    numbering/.style={circle, radius=1.5cm, fill=black, text=white, inner sep=0pt, outer sep=0pt}}
    \begin{tikzpicture}
    
 \node[matrix, column sep=-0.05cm, row sep=0.5cm] (m) {
            \node (pic1) {\includegraphics[width=0.31\linewidth]{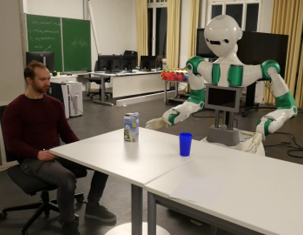}}; & \node (pic2) {\includegraphics[width=0.31\linewidth]{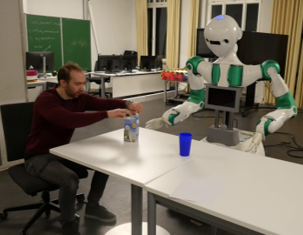}}; & \node (pic3) {\includegraphics[width=0.31\linewidth]{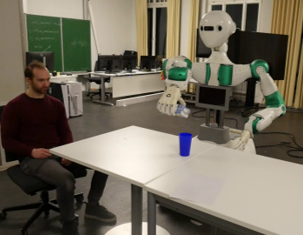}}; \\
        };
        \node[subcaption] [below of=pic1] {\scriptsize{Affordance Reasoning}};
        \node[subcaption] [below of=pic2] {\scriptsize{Planning \& Cooperation }};
        \node[subcaption] [below of=pic3] {\scriptsize{Plan Execution}};
        \node at (-2.8,1.4) [rectangle callout] (c1) [draw,fill=white, align=left, rounded corners=2, callout relative pointer={(-0.5cm,-0.5cm)}] {\scriptsize{I want a cup of milk.}};
        \node at (-0.75,1.4) [rectangle callout] (c2) [draw,fill=white, align=left, rounded corners=2, callout relative pointer={(0,-0.5cm)}] {\scriptsize{Sure.}};
        \node at (2, 1.4) [rectangle callout] (c3) [draw,fill=lightgray, align=left, callout relative pointer={(-0.7cm,-0.5cm)}] {\scriptsize{Can you open the milk for me?}};
        \node at (-1.4,-1.7) [rectangle callout] (c4) [draw,fill=lightgray, align=left, callout relative pointer={(-0.2cm,1cm)}] {\scriptsize{Can I use a cup instead of a glass?}};
        \node[numbering][above left=-1.5pt -1.5pt of c1.north west]{\tiny{\textbf{1}}};
        \node[numbering][above left=-1.5pt -1.5pt of c2.north west]{\tiny{\textbf{3}}};
        \node[numbering][above left=-1.5pt -1.5pt of c3.north west]{\tiny{\textbf{4}}};
        \node[numbering][above left=-1.5pt -1.5pt of c4.north west]{\tiny{\textbf{2}}};

    \end{tikzpicture}
    \caption{\armarDE solves the user task given in natural language by detecting the objects within the scene, reasoning about their affordances, planning how to solve the task including asking for help and finally executing the plan.}
\end{figure}
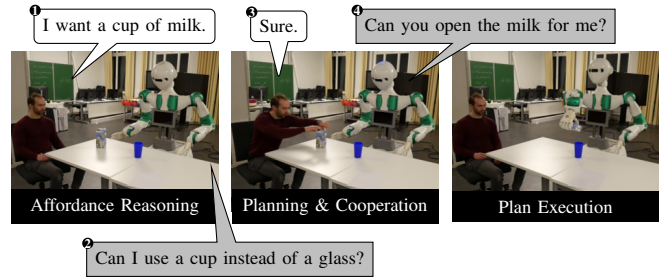

For instance, if a user requests a glass of milk but no glass is detected in the scene, \AutoGPTP proposes replacing the glass with a cup, ensuring task completion by considering alternative objects suitable for the task.
\added{When all task-relevant objects are accessible, \AutoGPTP can address the closed-world planning problem by extending the \LLMP approach with automated semantic and syntactic error correction and dynamic planning domain generation based on the agents' capabilities.}
Moreover, \AutoGPTP endows the robot with the ability to seek assistance from humans when \replaced{encountering problems while executing the actions needed to reach the goal}{executing an action needed to reach the goal surpasses the robot's capabilities}, such as requesting help with opening a milk box.

To formulate and execute plans effectively, \AutoGPTP relies on affordances, which represent the action possibilities that an object or environment offers to an agent \cite{gibson1977theory}. For instance, a knife affords cutting, grasping, or stirring. Leveraging the concept of affordances enables the dynamic deduction of viable actions within a given scene, facilitating the formation of a plan to achieve the user's objective. \added{Moreover, affordances allow for reasoning about how to substitute a missing object with suitable alternatives based on their functionality within the given task.}
%The robot must understand the interaction possibilities with the environment and the objects within it to form such plans. This is where Gibson's notion of affordances \cite{gibson1977theory} proves essential. Affordances describe a set of action possibilities an object or the environment presents to an agent. For example, a knife affords the possibility of cutting, grasping, or stirring. By utilizing affordances, it is possible to dynamically deduce feasible actions within a scene and form a plan to accomplish the user's objective. 

%Recently, \emph{Large Language Models} (LLMs) have become increasingly prominent in cognitive computing research and can function as effective zero-shot learners. However, directly transforming a natural language instruction into a plan can have several problems caused by the limited reasoning capabilities of LLMs \cite{valmeekam2023large} that state-of-the-art works like \cite{MichaelDosaycan} only partly address. 
%To deal with these limitations, \cite{liu2023llmp} suggest transforming natural language user instructions into a \emph{Planning Domain Definition Language} (\PDDL) problem.  In this manner, they transform the user's objective into an optimal plan that merges the benefits of integrating natural language tasks with the superior planning capacities of classical planners.

\AutoGPTP consists of two stages: the first involves perceiving the environment as a set of objects and extracting the scene affordances based on visual data. This is achieved by combining object detection and an \OAM (OAM), which describes the relations between object classes and the set of affordances associated with instances of those classes. Subsequently, in the second stage, task planning is conducted based on the established affordance-based scene representation and the user's specified goal. Here, \AutoGPTP utilizes an LLM to select tools that support generating a plan to accomplish the task.
\added{We quantitatively evaluate our approach in simulation using 180 scenarios with different goals to accomplish manipulation tasks like \pickplace, \handover, \pour, \chop, \heat, and \wipe. Additionally, we performed real-world validation experiments with a humanoid robot demonstrating a subset of these tasks.}

To summarize, the main contributions of this work are: 
\begin{enumerate*}[label=(\roman*)]
    \item a novel affordance-based scene representation combining object detection and automatic \OAM (OAM) using \ChatGPT
    \item  a task planning approach based on the established OAM and an LLM-based tool selection to generate plans, partial plans, explore and suggest alternatives in case of missing objects needed to achieve a task goal specified by the user in natural language,
    \item   
    \replaced{an evaluation of the approach in simulation using 150 scenarios with different tasks to accomplish like picking and placing, handover, pouring, chopping, heating, wiping, and sorting,}{an extension of the \LLMP planning approach with automated semantic and syntactic error correction and dynamic domain generation, and }
    \item real-world validation experiments with a humanoid robot demonstrating a subset of these tasks
\end{enumerate*}

The remainder of this work is structured as follows: First, we provide a comprehensive review of related work for both affordances and LLMs in planning tasks in \autoref{sec:related_work}. Then, in \autoref{sec:scene_representation}, we describe \removed{the general problem formulation and our proposed approach in Section IV}\added{our proposed scene representation followed by our planning approach \AutoGPTP in \autoref{sec:autogpt_p}}.
Subsequently, we discuss the results of our quantitative evaluation in simulation and our validation experiments on a humanoid robot.

\begin{table*}[h!]
    \begin{tabularx}{\textwidth}{|X|m{4.5cm}|m{3.5cm}|m{2.5cm}|m{1.0cm}|m{1.0cm}|} 
        \hline 
         Approach&  Planning Method&  \# Affordances &Substitutions&  Long-Horizon&  Novel Classes\\ \hline 
         \citet{lorken2008grounding} &  \PDDL + Planner &  2 (liftablibility \& pushability) &implicit&  no&  no\\ \hline 
         \citet{wang2013robotLearningAff}&  Reinforcement Learning&  1 (movability) & none & no &  yes\\ \hline 
         \citet{awaad2013affordance, Awaad_Kraetzschmar_Hertzberg_2014, awaad2015role} &  Hierarchical Task Network& (not directly listed) &explicit&  yes&  no\\ \hline
 \citet{moldovan2018relational}& Own Algorithm similar to Monte Carlo Search&  2 (tap and push)&none& no& no\\\hline
 \citet{chu2019toward, chu2019recognizing} & \PDDL + Planner&  7&implicit& no& yes\\\hline
 \citet{xu2022SGL}& \PDDL + Planner& 4 (grasp, cut, contain, support) &implicit& no& yes\\\hline
 \rowcolor{lightgray}Ours& \PDDL + Planner / Hybrid with LLM& 16 &explicit \& implicit& yes& no\\\hline
    
    \end{tabularx}
    \caption{Comparison to the state of the art in affordance-based planning.}
    \label{tab:comparison}
\end{table*}

\section{Related Work}
\label{sec:related_work}
\subsection{Affordances in Planning}

The use of affordances in \PDDL planning domains was initially proposed by \cite{lorken2008grounding} in a limited case study of using a crane-like robot to trigger switches with toy blocks. They do not distinguish between objects in their approach but only identify affordances so that specific objects cannot be selected. \PDDL is a generic planning language used to define planning domains and problems within those domains. When combined with a classical planner, this \PDDL goal allows generating a plan using a classical planner. The authors in \cite{chu2019toward, chu2019recognizing} expand the principle with a more sophisticated affordance segmentation approach to define the initial state of their \PDDL problem, which is then solved to generate a plan using affordances. Their experiments in three real-world manipulation tasks demonstrate the potential of combining detected affordances to design an affordance-based \PDDL domain. The work of \citet{xu2022SGL} trained an end-to-end model that learns to generate a \PDDL goal, consisting of (action, subject, object), from an input image and a natural language command, enabling the model to solve relatively simple tasks using an off-the-shelf task planner. All previously mentioned works use an affordance-based \PDDL domain, which implicitly allows them to replace objects with those of the same affordance.
In contrast to classical planning, the acquisition of relational affordances through probabilistic learning, used alongside a probabilistic planning algorithm that maximizes the likelihood of reaching the goal rather than minimizing plan length, was demonstrated in \cite{moldovan2018relational}. Relational affordances present a generalized affordance representation as a joint probability distribution over all objects, actions, and effects.
The authors of \cite{wang2013robotLearningAff} suggest using the (object, action, effect) relationship in task learning using reinforcement learning. Using actions only when the effect is relevant to achieving the goal reduces the search space and improves task learning in real-world navigation tasks.
The work of \cite{awaad2013affordance, Awaad_Kraetzschmar_Hertzberg_2014, awaad2015role} focuses mainly on affordance-based object replacement in planning tasks. Using a modified Hierarchical Task Network planning algorithm, they achieve flexible explicit object substitution based on functional affordances extracted by crawling dictionary definitions.

To the best of our knowledge, our work is the first to use affordance-based planning in everyday long-horizon tasks while employing implicit and explicit substitutions in planning. An overview of the related work can be found in \autoref{tab:comparison}. For the comparison, we define long-horizon tasks as tasks that need more than seven actions to be fulfilled, similar to \cite{MichaelDosaycan}. 

\subsection{Large Language Models in Task Planning} \label{sec:LLMsInPlanning}

\begin{figure}[b!]
\usetikzlibrary {arrows.meta}
\tikzset{ours/.style={fill=lightgray, text width=1cm}}
\tikzset{citations/.style={draw=none, text width=1.5cm}}
\tikzset{citationsbig/.style={draw=none, text width=3.5cm}}
\tikzset{citationsbiggest/.style={draw=none, text width=4.0cm}}
    \begin{tikzpicture}[
        level 1/.style={sibling distance=4cm, level distance=1.5cm},
        level 2/.style={sibling distance=3.5cm, level distance=1.7cm},
        level 3/.style={sibling distance=4cm, level distance=1.5cm},
        every node/.style={draw, text width=2.4cm, align=center}
    ]
        \node {LLMs for Planning}
            child { node {LLM as \\ Planner}
                child { node (Subtask) {Subtask Evaluation}
                }
                child { node {Autoregressive Plan Generation}
                    child { node (Full) {Full Plan Generation}}
                    child { node (Step_By_Step) {Step by Step Plan Generation}}
                }
            }
            child { node (LLM_With_Planner) {LLM with Planner}
            };

            \node [draw,ours] at (2.45,-3.2) (Ours) {\textbf{Ours} \emph{hybrid}};
            \node [citations] at (-3.75,-3.85) {\footnotesize{\cite{MichaelDosaycan, zhao2023large}}};
            \node [citationsbig] at (-2.25,-5.35) {\footnotesize{\cite{wu2023embodied, Wake2023, song2023llmplanner,rana2023sayplan,zhou2023isrllm,lin2023text2motion}}};
            \node [citationsbiggest] at (1.75,-5.35){\footnotesize{\cite{huang2022language, huang2022inner,Singh2023ProgPrompt,liang2023code,wu2023tidybot, wang2023describe, driess2023palme, baermann2024incremental}}};
            \node [citationsbig] at (2,-2.15){\footnotesize{\cite{xie2023translating,liu2023llmp,liu2024delta,guan2023leveraging,chen2023autotamp,ding2023integrating}}};
            \draw[line width=0.7mm, -{Latex[length=2mm]}] (2.45, -3.7) -- (2.45, -4.1);
            \draw[line width=0.7mm, -{Latex[length=2mm]}] (2.45, -2.7) -- (2.45, -2.3);
    \end{tikzpicture}
    \caption{A taxonomy of LLMs in planning tasks with the related work from this section referenced.}
    \label{fig:taxonomy}
\end{figure}
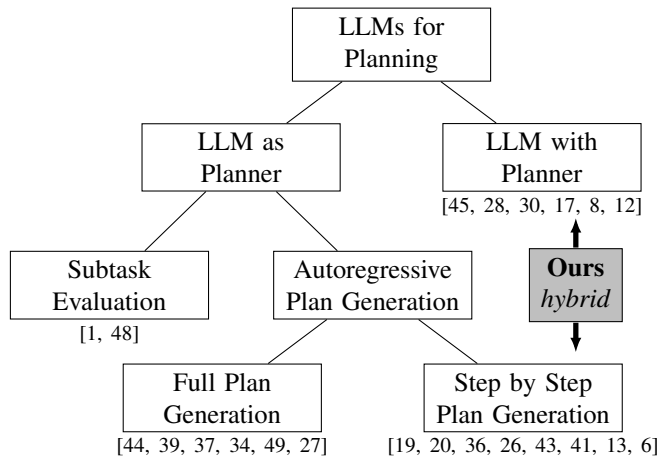

Recently, LLMs have demonstrated remarkable new capabilities and outperformed humans in various domains. However, their coherent reasoning skills remain somewhat lacking  \cite{valmeekam2023large}. Nevertheless, countless recent examples of using LLMs in robotic \replaced{applications}{task planning} exist.
According to the authors of \cite{sarkisyan23}, three general operation modes exist: subtask evaluation mode, full autoregressive plan generation, and step-by-step autoregressive plan generation. However, this classification is only sufficient when using the LLM as the planner. Recent studies, exemplified by \cite{liu2023llmp} and \cite{chen2023autotamp}, introduced a different paradigm where LLMs are used as symbolic goal generators in conjunction with a planner\added{, which we call \LLMwPlanner}. This section aims to distinguish between the different approaches and their respective categories, for which an overview is provided in \autoref{fig:taxonomy}.

\subsubsection{Subtask Evaluation}
In each step of plan generation, all possible actions are scored partly based on a probability provided by the LLM. The action with the best score is selected. One of the most well-known works on LLM-based task planning \emph{SayCan} \cite{MichaelDosaycan} utilizes a combination of a Reinforcement Learning-based affordance function and an LLM to predict the likelihood of an action. This affordance function describes the feasibility of an action in the given environment and differs from the previously described affordances. The plan is generated incrementally by selecting the action with the highest score resulting from multiplying the two functions.
\SayCan is expanded on in \cite{zhao2023large} by assessing the next-best action greedily and performing a tree search. They propose improving Monte Carlo Tree Search planning by incorporating an LLM that utilizes the context of the world state in the form of a common-sense heuristic policy.

\subsubsection{Full Autoregressive Plan Generation}
In this mode, the entire plan is generated based on the user-specified task.
The works of \citet{wu2023embodied} and \citet{Wake2023} provide a straightforward grounding method for the LLM by identifying objects in the scene, eliminating duplicates, and simply appending the object list to the prompt. The prompt instructs the LLM on what actions to generate and only permits using the provided objects when generating the plan.
\citet{song2023llmplanner} improve on that general idea by utilizing dynamic in-context example retrieval to enhance performance and enable the LLM to replan in the event of an error.
The authors of \cite{rana2023sayplan} reduce problem complexity by not giving the LLM all objects but filtering out irrelevant objects by traversing a 3D scene graph via collapsing and expanding notes before letting the LLM output a plan.
\citet{zhou2023isrllm} first translate the problem in natural language into a \PDDL domain and problem using the LLM. Then, based on this \PDDL domain, they let the LLM generate a plan, which the LLM itself can validate and a more precise external off-the-shelf \PDDL validator. The LLM can then use this feedback again to correct itself.
A hybrid approach between subtask evaluation and full plan generation is introduced in \cite{lin2023text2motion}, combined with a semantic checking of whether the goal condition is fulfilled. Their approach iteratively attempts to generate the entire plan. If the goal is not satisfied, a greedy step is taken to select the best next step according to a mixture of the LLM score and the skill feasibility.

\subsubsection{Step-By-Step Autoregressive Plan Generation}
In contrast to the previous mode, the plan is generated one action at a time, which allows for feedback from the execution of the action to improve the planning success rate.
\citet{huang2022language} addresses the issue of the LLM not being grounded in the actual scene and robot capabilities by introducing a two-step process: Firstly, a planning-LLM generates an ungrounded plan, which is then translated to the robot's abilities by a Translation-LLM. In a follow-up work~\cite{huang2022inner}, they first proposed generating the plan step-by-step with the LLM and continuously injecting feedback after each step to enhance the performance.
The works of \cite{Singh2023ProgPrompt,liang2023code,wu2023tidybot, baermann2024incremental} handle plan generation similarly but take advantage of the code generation capabilities of LLMs by letting it generate the plan as Python code. In Progprompt~\cite{Singh2023ProgPrompt}, feedback is injected with Python error messages to correct code that does not work due to syntactic or semantic errors. Tidybot~\cite{wu2023tidybot} additionally enables customization of preferences with a two-step method where the LLM initially identifies patterns from prior Python code and then generates new Python code with those patterns and supplementary instructions. \added{\cite{baermann2024incremental} proposes a system that additionally detects corrections of the plan in natural language by the human and incrementally learns to adapt plan generation in future tasks.}
\citet{wang2023describe} propose a 4-step approach: describe, explain, plan, and select, where the LLM is responsible for the describe and explain steps. In a feedback loop, the LLM iteratively generates plans that are executed until a failure occurs. Each time the plan fails, the descriptor describes the current state of the goal and the failed action in natural language. The explainer reasons why the plan failed, which is then fed to the planner, which returns a corrected plan. The selector prioritizes the actions in the plan to optimize its execution time.
\added{An embodied version of the LLM PaLM that can output plans in natural language based on multi-modal sentences is proposed in \cite{driess2023palme}. They embed visual data and the robot state as context for the user-specified task in natural language. Their approach PaLM-E allows embodied long-horizon task planning that can even handle adversarial disturbance.}

\subsubsection{LLM with Planner}
The authors of \cite{xie2023translating} introduce the concept of utilizing a pre-trained LLM to translate natural language commands into \PDDL goals. They argue that while LLMs are not adept at reasoning, which is essential for proper planning, they excel at translation. The conversion of natural language into a \PDDL goal can be seen as such a translation task. They demonstrate that LLMs are proficient in extracting goals from natural language in commonplace tasks. However, their accuracy diminishes with increasing task complexity. \LLMP~\cite{liu2023llmp} extends this idea beyond goal generation by generating the entire problem and using a classical planner to solve the task. They also improve the success rate by providing minimal examples for a similar goal state within the given domain. \added{An extension to \LLMP with scene graphs to generate the problem's initial state is presented in \cite{liu2024delta}. To reduce planning time, the authors decompose the overall goal into subgoals that can be solved more efficiently}.  \citet{guan2023leveraging} do not only use the LLM to generate the problem but also the domain itself using syntactic feedback to correct errors. Additionally, they propose a hybrid planning approach that uses plans generated by the LLM as a starting "heuristic" to speed up planning with a local search planner. The creators of AutoTAMP~\cite{chen2023autotamp} explore a similar direction, but instead of \PDDL, they use Signal Temporal Logic Syntax to define the goal state. Furthermore, they use an automatic syntactic and semantic checking loop that verbalizes the error to the LLM and tells it to correct it. In their case, a semantic error is defined as the resulting plan not being sufficient to solve the goal, which the LLM evaluates. In the work of \cite{ding2023integrating}, the LLM is used to expand the domain to be able to handle open worlds as \PDDL problems are specified under the closed-world assumption. This allows the system to handle situations not explicitly intended by the domain designer. Using a variety of prompts, they let the LLM generate augmentations of the \PDDL domain by defining new actions or changing allowed parameter types of actions like replacing a cup with a bowl containing water. This is functionally similar to affordance-based planning, except that in our case, the replacement of one object with another is implicitly given by their shared affordance. 

\added{Our work can be seen as a hybrid approach combining the \LLMwPlanner and the \emph{Step-By-Step Autoregressive Plan Generation} paradigms. We extend \LLMP by generating the initial state of the problem based on visual perception and the robot's working memory rather than natural language, which allows for a more dynamic plan generation. We also introduce automated syntactic and semantic self-correction for the generated \PDDL goal. Furthermore, \LLMP is limited to closed-world planning, which cannot handle missing objects. We overcome this limitation with \emph{Step-By-Step Autoregressive Plan Generation}, which iteratively updates the robot's memory by suggesting alternatives or exploring the scene until all necessary objects are found.}

\section{Affordance-based Scene Representation}
\label{sec:scene_representation}

\begin{figure*}
    \centering
    \includegraphics[width=1.0\textwidth]{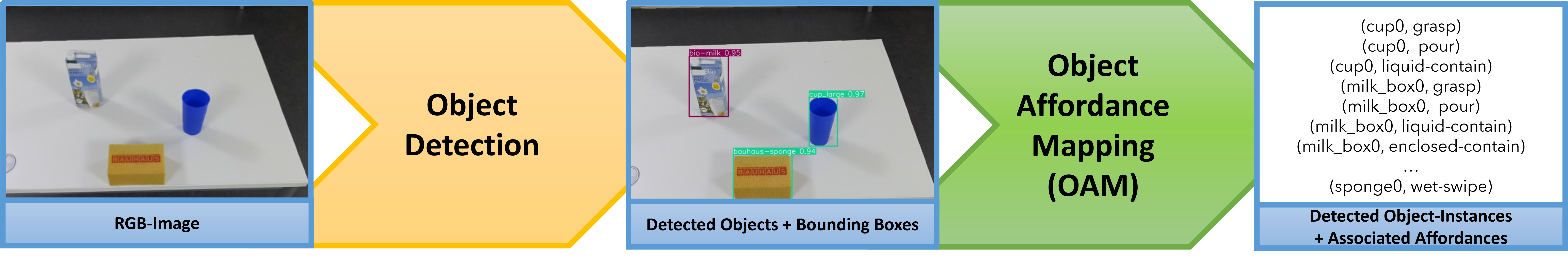} % Replace "example-image" with the actual filename of your image
    \caption{Overview of Object Affordance Detection(OAD). It uses an RGB image of a scene to detect the objects present in the scene. In the second step, the \OAM (OAM) maps the objects to their corresponding affordances.}
    \label{fig:OAD}
\end{figure*}

\replaced{The problem we address in this section is to obtain an affordance-based scene representation from an RGB image of the scene, which is used as grounding for the planner.}{The problem we address in this section is to extract an affordance-based scene representation from RGB images of the scene. Our task planning approach uses this representation to generate plans and reason about alternatives to missing objects. Affordances are particularly helpful in the planning context as they allow for actions in the planning domain to be defined by the functionality of the objects involved and not their class, allowing for a more generic planner \cite{lorken2008grounding}. Additionally, they provide useful information about how objects can be replaced with objects of the same functionality \cite{Awaad_Kraetzschmar_Hertzberg_2014}.} \removed{Since we focus on symbolic planning in this work, a simple representation of the objects and their affordances is sufficient.} To this end, we represent the scene $S$ symbolically as a set of object-affordance-pairs $p_i$, where each object has one or more affordances assigned to it as in \autoref{eq:scene}:
\begin{multline}
\label{eq:scene}
    S = \{p_1, \dots , p_n \} \text{, with}\\  p_i=(o_i,k_i,a_i, b_i) \in (\mathbb{O} \times \mathbb{N}_0 \times \mathbb{A} \times [0,1]^4),
\end{multline}

where $\mathbb{O}$ is the set of all object classes in the domain and $\mathbb{A}$ is the set of all possible affordances. $b_i$ represents the object's bounding box in normalized coordinates. Thus, the space of all scenes can be expressed as $\mathbb{S} = \mathcal{P}(\mathbb{O} \times \mathbb{N}_0 \times \mathbb{A} \times [0,1]^4).$ The problem of deriving this representation from the image of a scene, \ie Object Affordance Detection (OAD), can be formalized as in \autoref{eq:oad}:
\begin{equation}
\label{eq:oad}
    OAD:\mathbb{I} \rightarrow \mathbb{S}\\  
\end{equation}

% this paragraph could be omitted in case space is needed
Our definition of affordances aligns with the representationalist view, as discussed in \cite{affordances_robotics}. We do not use \citet{gibson1977theory} initial proposal, as we do not factor in the agent's capabilities or rely on visual features to identify affordances. Instead, we use a knowledge-based approach to extract affordances by detecting object classes.  
To this end, we detect the affordances of the object as a whole rather than identifying which specific parts of the object possess a particular affordance. We separate our approach into two distinct stages, as seen in \autoref{fig:OAD}. The first stage is object detection, and the second is the creation of an \OAM (OAM).
In the object detection stage, the goal, as defined in \autoref{eq:detection}, is to find a set of object instances $\hat{o} = (o, k) \in \mathbb{O} \times \mathbb{N}_0$ with their bounding boxes $b \in [0,1]^4$ in normalized coordinates given an image $I \in \mathbb{I}$. 
\begin{equation}
\label{eq:detection}
    ObjectDetection:\mathbb{I} \rightarrow \mathcal{P}(\mathbb{O} \times \mathbb{N}_0 \times [0,1]^4)
\end{equation}

The OAM associates object classes with the set of affordances assigned to the instances of those classes, leading to 
%, as can be seen in \autoref{eq:oam}.
\begin{equation}
\label{eq:oam}
    OAM:\mathbb{O} \rightarrow \mathcal{P}(\mathbb{A})
\end{equation}

The OAM can be generated offline and stored in a database \added{as presented in \autoref{sec:cgpt_oam}.} \replaced{which can then be loaded for online detection.}{During inference, the previously generated OAM can be used.}
The complete approach can be expressed by %\autoref{eq:oad_function}:
\begin{multline}
\label{eq:oad_function}
    OAD(I) = \{ (o,k,a,b)  \mid \\ a \in OAM(o), (o, k, b) \in \text{ObjectDetection}(I) \}
\end{multline}

\subsection{\OAM using \ChatGPT}
\label{sec:cgpt_oam}
LLMs have the ability to reproduce real-world knowledge when prompted in natural language. This is especially true for commonsense knowledge (\cite{Kandpal2022LargeLM}), including interaction possibilities with everyday objects. \replaced{We exploit this by employing different strategies for querying an OAM from the LLM-based chatbot \ChatGPT, which we describe in the following.}{We exploit the knowledge reproduction capabilities by querying the LLM with simple questions that do not involve complex reasoning or context understanding.} We omit the formatting instructions for the LLM in the prompts for brevity.
\begin{itemize}
    \item 
\textbf{List-Affordances:} In this strategy, we iterate over all objects we want to get the affordances for and ask the LLM which affordances the object has. As context, we give the LLM a list of affordances with a short description for each affordance. 
This has the advantage of using only a few tokens and is relatively fast. However, it lacks accuracy, possibly because the descriptions of affordances are sometimes unclear.

\item \textbf{Yes/No-Questions:} In this strategy, we define a prompt formulated as a yes-no question for each affordance. We then query \ChatGPT to answer the question with only yes or no without an explanation.
In contrast to the first strategy, we can describe precisely what an affordance means. This improves accuracy as \ChatGPT only needs to generate the answer to a binary question. However, we need significantly more tokens and time for the task.

\item \textbf{Yes/No-Questions + Logical Combinations}: Early experiments showed that \ChatGPT cannot handle logical combinations of questions very well. Therefore, in this strategy, queries contain multiple questions divided into atomic prompts, only containing one question each.
This approach has the advantage of being more accurate than all other approaches. Still, it has the disadvantage of consuming even more tokens than the yes/no questions, as it often requires multiple questions per affordance.

\end{itemize}

\begin{figure*}[h!]
    \centering
    \includegraphics[width=1.0\textwidth]{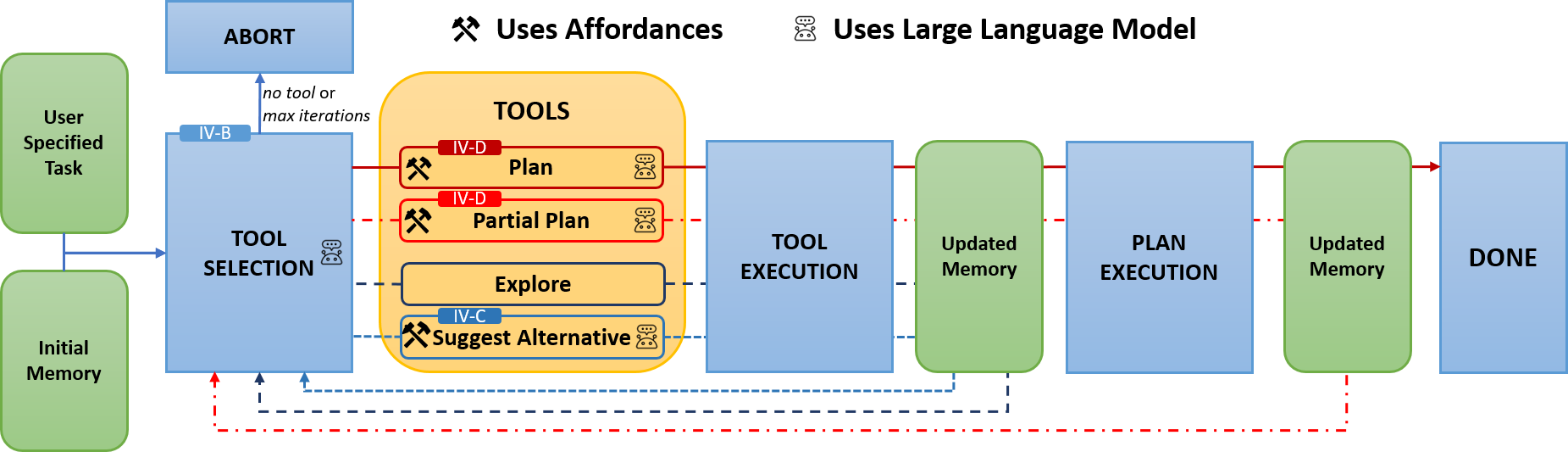}
    \caption{Overview of the \AutoGPTP feedback loop presented in \autoref{sec:loop}. \replaced{The left column shows the tools that the LLM can select. The center column shows the main feedback loop as a statechart with conditional transitions in brackets. The right column shows the contents of the memory that is updated with tool execution and used to generate context for the LLM.}{Green boxes symbolize inputs and outputs, while blue boxes symbolize discrete steps of the process. The tool selection process chooses one of the tools in the yellow \emph{Tools} box. The numbers on top of the boxes show in which section the aspect of the work is explained.} }
    \label{fig:approach}
\end{figure*}
\section{AutoGPT+P}
\label{sec:autogpt_p}
To generate plans from a user command, \AutoGPTP uses a tool-based architecture. \replaced{Given the scene's current state, the main planning loop queries an LLM to decide which tool should be selected next. Therefore, the loop generates a context from the robot's memory, which includes objects in the scene and their relations, as well as prior knowledge about the environment.}{The tools are used to iteratively update the robot's memory, which contains the affordance-based scene representation until a final plan is found.} A general overview of our approach can be seen in \autoref{fig:approach}.

\subsection{Problem Formulation}
\label{sec:problem_planning}

In the following, let $\mathbb{R}_S$ be the set of all possible object relations in the scene $S$ and $\Lambda$ be the space of natural language. The overall planning task can be specified as given a scene description $S \in \mathbb{S}$, object relations $R \in \mathbb{R}_S$, explorable locations $L$, and a task in natural language $\lambda \in \Lambda$, the system should return an action sequence or plan $P = (\alpha_1, \dots, \alpha_n)$ that fulfills the task.
An action $\alpha_i \in A$ is defined as the executed capability $c$ by the agent $\pi$ with the arguments $\rho = (\rho_1, \dots, \rho_n)$. Here, $A$ refers to the set of all available actions, and a capability defines the symbolic parameters of an action with their logical preconditions and effects. Each agent has a set of capabilities $C_\pi$ that are dynamically loaded at run-time and are derived from the available skills, which are the programs for low-level action execution on the robot. $S$ can be updated during the process by exploring a location $l \in L$ and adding the object-affordance-pairs $\hat{p} = OAD(I)$ to $S$ with the image $I$ taken at $l$.

Two relevant sub-tasks during planning are the \textit{closed-world planning} problem and the \textit{alternative suggestion} problem.
The closed-world planning based on tasks in natural language can be described as follows: Given the fixed scene representation $S \in \mathbb{S}$, object relations $R \in \mathbb{R}_S$, and the user-specified task $\lambda \in \Lambda$, we need to generate a plan $P = (\alpha_1, \dots, \alpha_n)$ that fulfills the given task. This can be written as  % as given in \autoref{eq:llmp}.
\begin{equation}
\label{eq:llmp}
    ClosedWorldPlanning: (\Lambda \times \mathbb{S} \times \mathcal{P}(\mathbb{R}_S)) \rightarrow A^\mathbb{N}
\end{equation}

An alternative suggestion is the problem of suggesting an alternative object $alt \in O$, where $O$ is the set of object classes present in the scene, given a user-specified task $\lambda$ in natural language and a missing object class $o \in \mathbb{O}$ needed to fulfill that task. This can be written as 
\begin{equation}
\label{eq:alternative}
    AlternativeSuggestion: \Lambda \times \mathbb{O} \rightarrow O
\end{equation}

\subsection{AutoGPT+P Feedback Loop}
\label{sec:loop}
\AutoGPTP is a hybrid planning approach that combines two planning paradigms introduced in \autoref{sec:LLMsInPlanning}: \emph{Step-By-Step Autoregressive Plan Generation} for tool selection and an \LLMwPlanner in the \emph{Plan Tool} (\autoref{sec:llmp}). \added{The motivation behind this design is to fill the robot's memory with information using the main feedback loop as specified in \autoref{algo:tool_selection_loop} until the \emph{closed-world planning problem} is solvable with the \emph{Plan Tool}, thus making the planning process more versatile.} 
\removed{The tool selection process was inspired by the GitHub project AutoGPT, hence the name AutoGPT+P.}

The tool selection is the central part of the main feedback loop of \AutoGPTP.
It can be specified as %in \autoref{eq:tool_selection}.
\begin{equation}
\label{eq:tool_selection}
    ToolSelection: \Lambda \times \mathbb{M} \rightarrow T,
\end{equation}
where $\mathbb{M}$ is the space of memory configurations and $T = \{t_1, \dots t_n\}$ is the set of tools. So based on the user prompt $\lambda \in \Lambda$ and the current memory state $M \in \mathbb{M}$, the tool selection returns a tool $t \in T$.
The memory $M = (S, R, L, l_\Pi, \hat{\lambda}, Alt, \replaced{p}{P})$ consists of an affordance-based scene representation $S$, a set of object relations $R$, locations $L$, current agent locations $l_\Pi$, instruction history $\hat{\lambda}$, known alternatives $Alt \in (\mathbb{O} \times O)$ and most recent plan $\replaced{p}{P}$. With $M$ expressed in natural language, the LLM chooses from the following tools:

\begin{itemize}
    \item \textbf{Plan}: solves the problem expressed in \autoref{eq:llmp}
    \item \textbf{Partial Plan}: solves the problem expressed in \autoref{eq:llmp} in the best way possible with the restrictions of $S$.
    \item \textbf{Suggest Alternative}: solves the problem expressed in \autoref{eq:alternative}
   \item \textbf{Explore}: move the robot to an unexplored location $l \in L$, extracts the object-affodance-pairs from the camera image and updates the scene representation: $S \leftarrow S \cup OAD(I)$.
\end{itemize}

After $t$ is selected, it is executed to update the memory $M$. The generated plan is executed if either the \emph{Plan} or \emph{Partial Plan Tool} is selected. \added{The plan execution assumes all actions are executed successfully and does not replan on failure.} These steps are repeated until a final plan is generated via the \emph{Plan Tool}, or no tool is selected, which is interpreted as a failure.
\removed{The entire loop is given in Algorithm 1.}
\begin{algorithm}[t]
    \KwIn{Memory $M\added{= (S, R, L, l_\Pi, \hat{\lambda}, Alt, P)}$, User-specified task $\lambda$}
    \While{$\neg final$}
    {
        $t \leftarrow ToolSelection(\lambda,M)$ \\
        \If{t}
        {
            $M, final \leftarrow ExecuteTool(t, M)$\\
        }
        \Else
        {
            \Return
        }
        \If{$M.p$}
        {
            $executePlan(M.\replaced{p}{P})$\\
        }
    }
    \caption{AutoGPT+P Feedback Loop}
	\label{algo:tool_selection_loop}
\end{algorithm}

\subsection{Affordance-Based Alternative Suggestion}
\label{sec:alternative}
\added{If an explicitly requested object cannot be found within the scene, our system should reason whether another object in the scene can replace it.}
One of the reasons for choosing an affordance-based scene representation was that affordances allow us to reason about the functionality of an object. We leverage this reasoning for the alternative suggestion task defined in \autoref{eq:alternative}. \removed{If an object that the user explicitly requests is missing in the scene, we can suggest an alternative based on the object affordance using an LLM.}

Our method uses a handcrafted Chain-of-Thought process \cite{wei2023chainofthought} consisting of two main steps, detailed in \autoref{algo:alternative_suggestion}. First, we query the LLM which of the affordances of the missing object class $aff_{m}$ are relevant to the task $\lambda$ specified by the user. We can now filter out all objects in the scene that do not have all these affordances and get the most relevant, which we heuristically assume is the rarest affordance in the scene $a*$. Now, we query the LLM to find out which of the objects is the most similar to the missing object concerning this affordance.
If no objects have all the affordances, or the LLM returns an object not present in the scene, the fallback strategy is to query the LLM to suggest the best replacement for the missing object without explicit affordance reasoning. \added{Our evaluation in \autoref{sec:eval_plan_tool} demonstrates that the use of affordances substantially enhances the accuracy of the substitution.}

\subsection{Affordance-Based Planning using an LLM with Planner}
\label{sec:llmp}

\begin{algorithm}[t]
	\KwOut{Suggested Alternative $alt \in O$}
    \KwData{\OAM $OAM$, Large Language Model $LLM$}
    $aff_{m} \leftarrow OAM(m)$\\
    $aff_{rel} \leftarrow LLM.askForAffordances(m, aff_{m}, \lambda)$\\
    $O_{filtered} \leftarrow \{ o* \in O \mid OAM(o*) \subseteq aff_{m} \}$\\
    \If{$O_{filtered} = \emptyset$}
    {
         \Return $LLM.askForObjectDirect(m, O, i)$
    }
    $a* \leftarrow rarestAffordance(aff_{rel}, O)$\\
    $alt \leftarrow LLM.askForObject(m, a*, O_{filtered})$\\
    \If{$!(alt \in O_{filtered})$}
    {
         \Return $LLM.askForObjectDirect(m, O, \lambda)$
    }
    \Return $alt$\\
    \caption{Alternative Suggestion}
	\label{algo:alternative_suggestion}
\end{algorithm}

\replaced{Inspired by \cite{liu2023llmp}, this aspect of our approach aims to create a system mapping user-specified tasks in natural language into a sequence of parameterized actions $\alpha_i$ under a closed-world assumption as defined in Equation \ref{eq:llmp}. A key difference to the approach of \cite{liu2023llmp} is that we only generate the goal state from natural language and not the entire problem formulation as we derive the initial state from the given scene representation.}{This component of our approach maps user-specified tasks in natural language to a sequence of parameterized actions $\alpha_i$ under a closed-world assumption as defined in \autoref{eq:llmp}. It is an extension of \LLMP \cite{liu2023llmp}, with the key differences being that our \emph{Plan Tool}
\begin{enumerate*}[label=(\roman*)]
    \item generates the initial state of the \PDDL problem from our affordance-based scene representation and not from natural language,
    \item dynamically generates the \PDDL domain based on the capabilities of the agents and the OAM, and
    \item checks for the semantic and syntactic correctness of the generated goal and lets the LLM correct its own errors.
\end{enumerate*}
}

\begin{figure*}
    \centering
    \includegraphics[width=1.0\textwidth]{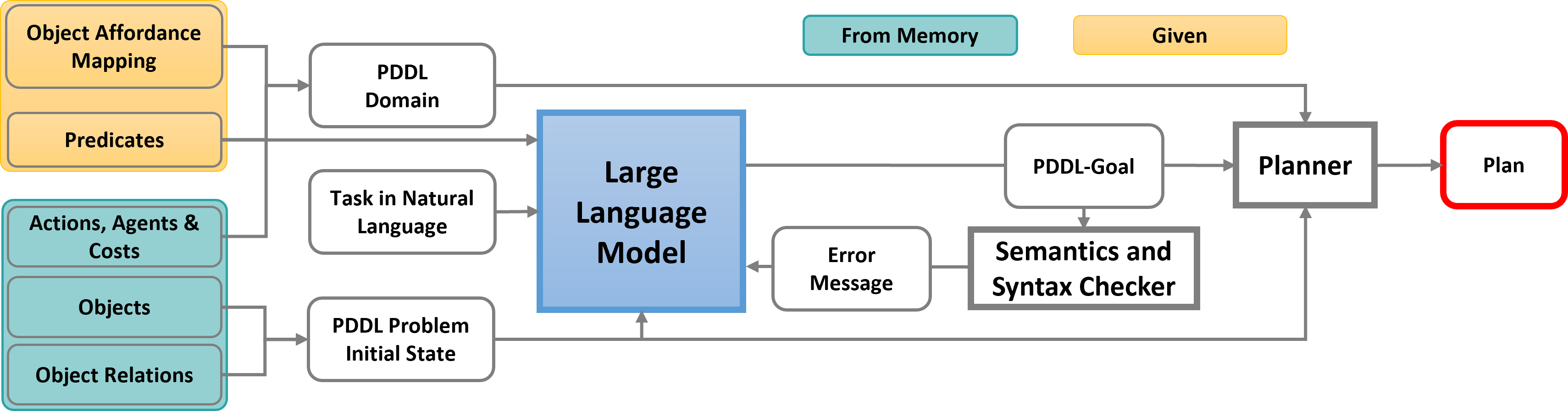} % Replace "example-image" with the actual filename of your image
    \caption{Overview of the Planning Tool. Rounded boxes represent the input and the output of the components that are represented as rectangles. }
    \label{fig:llmp}
\end{figure*}

Similar to \LLMP, we let the LLM generate the desired goal state $\Gamma$ in \PDDL syntax from the user-specified task $\lambda$. Therefore, we need to generate a \PDDL domain $\Delta = (\Theta, \Phi, A)$ and problem without the desired goal state $\hat{\Xi} = (\omega, \iota)$ as a reference for the LLM as explained in \autoref{sec:domain}. The generated goal $\Gamma$ is then checked for semantic and syntactic correctness as seen in \autoref{sec:oracle}. If there is an error within the goal, the LLM is fed an error message $error \in \Lambda$ and queried to correct the goal state. If the goal is correct, a classical planner is invoked with the generated domain and problem. This process is repeated until a plan is found or the maximum number of iterations has been reached. It is more formally described in \autoref{algo:llmp} and visualized in \autoref{fig:llmp}.

We use the method for both the \emph{Plan} and \emph{Partial Plan Tool}. The only difference between the tools is the prompt used to query the goal state from the LLM, which explicitly allows for an incomplete goal state.
A significant advantage of this method \added{compared to \LLMasPlanner} is that if the symbolic goal representation accurately represents the given user-specified task, the generated plan will be optimal regarding the number of actions.
\\

\begin{algorithm}[t]
	\KwIn{User-specified task $\lambda$, Scene $S$, Object Relations $R$, Agent Locations $l_\Pi$ Capability Sets of Agents $C_\Pi$, maximum loops $n$}
	\KwOut{Plan $P$}
    \KwData{Large Language Model $LLM$, Predefined Predicates $\Phi$, Semantic Conditions $\Sigma$, Planner $planner$}
    $\Delta \leftarrow createDomain(S, C_\Pi, \Phi)$\\
    $\hat{\Xi} \leftarrow createProblemInitialState(\Delta, S, R, l_\Pi)$\\
    $loops \leftarrow 0$\\
    \While{$!P \And loops < n$}
    {
        $loops \leftarrow loops + 1$\\
        \If{error}
        {
            $\Gamma \leftarrow LLM.correctGoal(error)$\\
        }
        \Else
        {
            $\Gamma \leftarrow LLM.askForGoal(\Phi, \hat{\Xi})$ \\
        }
        $\Xi \leftarrow (\omega,\iota,\Gamma)$\\
        $error \leftarrow checkSyntax(\Delta, \Xi)$\\
        $error \leftarrow checkSemantics(\Gamma, \Sigma)$\\
        \If{!error}
        {
            $P = planner.plan(\Delta,\Xi)$\\
        }
    }
    \Return $P$
    \caption{Planning with Self-Correction with external feedback}
	\label{algo:llmp}
\end{algorithm}

\subsubsection{Dynamic Generation of the Affordance-based Domain and Problem}
\label{sec:domain}
For the LLM to generate the desired goal state based on the user-specified task, it needs a \PDDL domain $\Delta$ and problem $\Xi$ without the goal state as context. 

In \PDDL, types are defined by listing all subtypes of a given type. Our domain has three top-level types: object, agent, and location. As it should be possible to navigate to another agent too, for example, to hand over an object, the agent is a subtype of location. Let $sub(\theta)$ define \added{the set} of subtypes of a given type.
To build the type hierarchy, we first need to declare all affordances as object subtypes, so $sub(object) = \mathbb{A}$. \replaced{Then we reverse the OAM so it maps from affordances to all object classes that are present in the scene $O$ that also have that affordance.}{Afterwards, we need to map all object classes that have a given affordance to the subtype of that affordance.} So for all affordances $a \in \mathbb{A}$ \autoref{eq:affordance_types} holds.
\begin{equation}
    \label{eq:affordance_types}
    sub(a) = \{o \mid a \in OAM(o), o \in O\} 
\end{equation}
To model the different capabilities of agents, we need to define a type for each capability available, so $sub(agent) = C_\Pi$. Additionally, we make all agents that have a capability $c$ subtypes of that capability type, so for all capabilities $c \in C_\Pi$ \autoref{eq:capability_types} holds.
\begin{equation}
\label{eq:capability_types}
    sub(c) = \{\pi \mid c \in C_\pi\ \}
\end{equation}
To allow for human-robot collaboration, we need to define the different agent types, robot, and human, and dynamically assign costs to them based on user preference.

The actions can be derived directly from the capabilities by adding an agent of the corresponding capability type to the parameters and adding an action effect to increase the total costs based on the agent type. By associating specific costs for agents in the initial state specification of the \PDDL problem, we can influence the participation of each agent. For example, by setting the cost of a human to 1000 and of the robot to 1, the robot will execute all actions that it can perform with its capabilities and will only ask the human for help if there is no other possibility.

To define the problem's initial state, we add each object instance with its type to the problem's object definition. The initial state can be directly derived from $R$; only the current agent locations are given by $l_\pi$. Finally, the goal state is queried from the LLM to complete the problem.

The advantage of using affordances in our domain is that we only need to define one logical action for all objects with which the action can be performed. Without affordances, we would need a \emph{place} action for all combinations of objects on which other objects can be placed. This would make the domain far more complex and thus increase the search time of the planner.
\\

\subsubsection{Self-Correction with External Feedback}
\label{sec:oracle}
The work of \cite{gou2023critic} demonstrated that conversational agents using LLMs can correct themselves when an external program gives an expressive error message. We leverage this capability by detecting syntactic and semantic errors within the goal state \added{and thereby help ground the LLM}. A syntactic error can be a wrong use of parenthesis, non-existent predicates, non-existing objects, or predicates with objects of the wrong type or quantity. When parsing the goal state, those errors can be easily checked by matching the predicate names and object names and types with those of the domain and initial state.

\replaced{Semantic errors refer to the logical feasibility of multiple predicates being true simultaneously.}{We define semantic errors as the occurrence of multiple predicates that cannot be true at the same time in a real scene.} This contrasts to \cite{chen2023autotamp}, where a semantic error is defined as an action sequence that does not fulfill the user-specified task according to the LLM. For example, the object apple cannot be on the table and the counter at the same time, so the goal state {\footnotesize \texttt{and (on apple table) (on apple counter)}} is semantically incorrect. We can express those semantic conditions $\Sigma_\Delta$  using predicate logic and check whether a goal matches a semantic condition in the logic programming language \emph{Prolog}\cite{prolog}, which is an implementation of first-order predicate logic. In the following definition, let the disjunctive normal form (DNF) of $\Gamma$ be $DNF(\Gamma)=OR(\gamma_1, \dots,\gamma_n) \text{ with } \gamma_i = AND(\varphi_{i,1}, \dots, \varphi_{i,m_i})$. We define a goal state $\Gamma$ to be sufficient to the set of semantic conditions of the domain $\Sigma_\Delta$ if there exists $\gamma_i$ where $\gamma_i$ is sufficient to all conditions in $\Sigma_\Delta$.

Therefore, \added{in the semantic error check specified in \autoref{algo:semantic_error_check}}, we transform the goal state $\Gamma$ into its DNF and map all sub-states $\gamma_i$ to Prolog predicates. We then evaluate all semantic conditions for these predicates. If no sub-state matches all semantic conditions, we return the manually specified error message of the condition of the sub-state with the fewest errors to the LLM to correct itself.

\begin{algorithm}[t]
	\KwIn{Generated Goal State $\Gamma$, Set of semantic conditions $\Sigma$}
	\KwOut{Unfulfilled Condition $\sigma$}
    $dnf \leftarrow transformToDNF(\Gamma)$\\
    $best \leftarrow \emptyset$\\
    \For{$\gamma \in dnf$}
    {
        $failed \leftarrow \emptyset$\\
        \For{$\sigma \in \Sigma$} 
        {
            \If{$!checkCondition(\gamma, \sigma)$}
            {
                $failed \leftarrow failed \cup \sigma$\\
            }
        }
        \If{$|best| > |failed|$}
        {
            $best \leftarrow failed$
        }
    }
    \Return $getMostGeneral(best)$\\
    \caption{Semantic Error Check}
	\label{algo:semantic_error_check}
\end{algorithm}

\begin{comment}
\subsection{Plan Execution}

For plan execution, as in \autoref{algo:plan_execution}, the generated symbolic action sequence $P = (\alpha_1, \dots, \alpha_n)$ needs to be mapped to the executable skills of the agents. Therefore, each symbolic capability $c$ has an executable skill $s$ mapped to it in the agentSkillMapping (ASM) and procedures that determine how to map the symbolic representations of the parameters to the actual objects in the robot's memory. This is needed as only the localized instances with a pose can be interacted with. Additionally, for human-robot collaboration, the agents have to be separated by their type, so the skill mapping is only done for agents with the robot type. For human agents, the actions they need to perform are verbalized and need to be manually confirmed by the human for the plan execution to continue. Note that we do not verify the successful execution of a skill.

\begin{algorithm}[h!]
	\KwIn{Symbolic Action Sequence $P$}
    \KwData{agentSkillMapping $ASM$}
    \For{$(\pi, c, \rho) \in P$}
    {
        \If{$type(\pi) = robot$}
        {
            $skill \leftarrow ASM(c,\pi)$\\
            $args \leftarrow getArgsForSkill(c, skill,\rho)$\\
            $executeSkill(skill, args)$\\
        }
        \Else
        {
            $verbalize(c,\rho)$\\
        }
    }
    \caption{Plan Execution}
	\label{algo:plan_execution}
\end{algorithm}
\end{comment}

\section{Evaluation and Validation}

We first evaluate the performance of the automated OAM on our proposed affordances. Then, we assess the success rate of our \emph{Suggest Alternative Tool} against a naive alternative suggestion. Furthermore, we compare the \emph{Plan Tool} on its own against \SayCan on the \SayCan instruction set and our own evaluation set before evaluating the whole AutoGPT+P system with scenarios focused on tool selection. In this evaluation, \GPTIII refers to the \texttt{GPT-3.5-turbo-0613} model, and \GPTIV refers to the \texttt{GPT-4-0613} model accessed by the OpenAI API.

The quantitative evaluation is conducted via simulation wherein a scene is represented through symbolic object relations, and actions are executed by applying their respective action effects to the scene. For evaluating the \emph{Plan Tool}, all objects in the scene are known from the beginning, unlike \AutoGPTP, where this is not always the case. Exploration was simulated by changing the robot's location and adding all objects designated to the explored location to the robot's memory. We use \emph{Fast Downward} \cite{Helmert_2006} as the planner with a time limit of 300 seconds.

We designed a collection of evaluation scenarios, each consisting of the user's task, the formal goal state to be achieved, and the specifications of the scene, including the objects, relations, and locations. The primary evaluation criterion was whether the generated plan achieves the objective that meets the desired goal state. This can be verified by simulating the plan's actions using Prolog, transforming the goal into its DNF, and assessing whether the set of literals that comprise a sub-state of the DNF is a subset of the literals that describe the scene state after executing the plan.

\subsection{Object Affordance Mapping using ChatGPT}

To evaluate the OAM, the relevant metrics\footnote{All these metrics fall within the range of 0 to 1, with higher values indicating better performance.} are precision (prec), recall (rec), and F1-score (F1) with
\begin{equation*}
    \text{prec} = \frac{TP}{TP + FP} \; , \; \text{rec} = \frac{TP}{TP + FN}\; ,\; \text{F1} = 2 \times \frac{\text{prec} \times \text{rec}}{\text{prec} + \text{rec}}
\end{equation*}
where in our case
\begin{itemize}
    \item TP is the number of true positives, so object-affordance-pairs (OAP) that are both in the ground truth (GT) and were detected
    \item FP is the number of false positives, so OAPs that were detected but not the GT
    \item FN is the number of false negatives, which is the number of OAPs that are in GT but were not detected
\end{itemize}

An independent "training set" of 30 object classes was used to optimize the prompts. The evaluation involved a test set comprising 70 object classes, each labeled with their respective affordances. These were also used to evaluate the \emph{Plan} and \emph{Suggest Alternative Tool}. We examined the metrics for different affordance extraction methods listed in \autoref{sec:cgpt_oam} with our 40 proposed affordances that can be seen in the appendix.

\begin{table}[H]
    \centering
    \vspace*{\fill}
    \noindent\makebox[\linewidth]{%
    \begin{tabular}{|c|c c c|c c c|c c c|} 
    \hline 
     \multirow{2}{*}{GPT}& \multicolumn{3}{c|}{List-Affordances}  &  \multicolumn{3}{c|}{Yes/No} & \multicolumn{3}{c|}{Logical} \\
         & prec &rec &F1 & prec &rec &F1 & prec &rec &F1 \\
         \hline
         3& 0.31 & 0.49 & 0.38 & 0.70 & 0.78 & 0.74 & 0.78 & 0.85 & 0.81\\ \hline 
         4& 0.59 & 0.67 & 0.62 & 0.78 & \textbf{0.95} & 0.86 & \textbf{0.87} & 0.91 & \textbf{0.89}\\ 
         \hline
    \end{tabular}%
    }
    \caption{Comparison of \ChatGPT OAM methods on our proposed set of affordances for affordance-based planning with the best values for precision, recall, and F1-score in bold}
    \label{tab:oam_eval_proposed}
\end{table}

As the results in \autoref{tab:oam_eval_proposed} indicate, \GPTIV outperforms \GPTIII in most cases. The data suggests that the most effective method is the combination of yes/no questions and logic.
\added{Despite achieving a high level of accuracy in affordance detection, the uncertainty in affordance estimation is a factor that should be considered in future work.}

\subsection{Suggest Alternative Tool}
For the \emph{Suggest Alternative Tool}, we compare our approach with a naive alternative suggestion approach. This approach asks the LLM to determine which object from the scene can best replace the missing object without any further reasoning. We evaluate the performance of both methods using 30 predefined scenarios. Each scenario includes the missing object, the user-specified task, the objects in the scene, and a list of allowed alternative objects. The task is considered accomplished if the method provides one of the permitted alternatives. 

We use 30 scenarios with three difficulty levels, each based on the number of objects present in the scene. The first level is simple and involves five objects. The medium level has twenty objects, whereas the complex level has 70 objects, with one missing. Our rationale behind this setup is that as the number of objects in a scene increases, it becomes more challenging to identify the missing object accurately.
Our approach and the naive approach were assessed using \GPTIII and \GPTIV, and the results are available in \autoref{tab:evaluation_alternative}.

\begin{table}
    \centering
    \noindent\makebox[\linewidth]{%
    \begin{tabular}{|c|c c|c c|} 
    \hline 
         \multirow{2}{*}{} &  \multicolumn{2}{c|}{\GPTIII} & \multicolumn{2}{c|}{\GPTIV}\\
         & Naive & Ours & Naive & Ours\\ \hline 
         simple&  0.73&  0.87&  \textbf{0.90}& \textbf{0.90}\\ \hline 
         medium&  0.63&  \textbf{0.90}&  0.70& 0.83\\ \hline 
         complex&  0.33&  0.80&  0.67& \textbf{0.80}\\ \hline
    \end{tabular}%
    }
    \caption{Comparison of the success rate of our \emph{Suggest Alternative Tool} with a naive approach. The best values for the success rate are in bold.}
    \label{tab:evaluation_alternative}
\end{table}

We found that as the number of objects in the scene increases, all approaches experience decreased accuracy. However, compared to the naive approach, which experiences a significant drop in accuracy from 0.73 to 0.33 for \GPTIII and from 0.9 to 0.67 for \GPTIV, our approach has only a slight drop in accuracy from 0.9 to 0.8 and from 0.87 to 0.8, respectively. In addition, unlike the naive approach, there is only a slight difference in accuracy between \GPTIII and \GPTIV when using our approach. This could be because the LLM is guided through the replacement process by a directed Chain-of-Thought process, eliminating incorrect answers.

\subsection{Plan Tool} \label{sec:eval_plan_tool}

Our \emph{Plan Tool} was assessed using two sets of scenarios. The first set comprised scenarios from \SayCan \cite{MichaelDosaycan}, which was utilized to draw comparisons between our method and \emph{SayCan}, a state-of-the-art planning approach using LLMs. We created the second set of scenarios to find the limitations of the LLM's reasoning capabilities for understanding the user's intentions. Therefore, we created five subsets of scenarios, each of them containing 30 prompts with a wide variety of goal tasks from cutting, heating, cleaning, pouring, opening, or moving objects.
The \emph{Simple Task} and \emph{Simple Goal} subsets contain simple user requests using either a verb to express the goal or the goal in the form of a state. \added{\emph{Complex Scene} contains task similar to \emph{Simple Goal} but in scenes with 100 instead of around 30 objects like the scenes in all other subsets.} \emph{Complex Goal} increases complexity compared to \emph{Simple Goal} by logically connecting the subgoals with phrases like \emph{"and"}, \emph{"or"}, \emph{"if"}, etc. The other two sets \emph{Knowledge} and \emph{Implicit} contain more difficult-to-understand tasks. \emph{Knowledge} requires commonsense knowledge to derive the goal state, while the \emph{Implicit} set does not directly contain a task but more implicitly phrased user intentions like "I am thirsty".

\begin{comment}

\begin{table}
    \begin{tabular}{|M{2.2cm}|M{1.65cm}|M{1.6cm}|M{1.6cm}|}\hline
         Instruction Family&  \SayCan (plan)& Ours (\GPTIII)& Ours (\GPTIV)\\ \hline 
         NL Primitive &  0.93 & 0.73& \textbf{1.00}\\ \hline 
         NL Verb &  0.60 & 0.33& \textbf{1.00}\\ \hline 
         NL Noun &  0.93 & 0.40 & \textbf{1.00}\\ \hline 
         Structured&  \textbf{0.93} & 0.13& \textbf{0.93}\\ \hline 
         Embodiment&  0.64 & 0.64 & \textbf{1.00}\\ \hline 
         Crowd-Sourced&  0.73 & 0.33 & \textbf{0.93}\\ \hline 
         Long-Horizon&  0.73 & 0.33 & \textbf{1.00}\\ \hline 
         Drawer&  \textbf{1.00}& 0.33 & \textbf{1.00}\\ \hline
         \rowcolor{lightgray} Average& 0.81 & 0.40&\textbf{0.98}\\ \hline
    \end{tabular}
    \caption{Comparison of the success rates for planning from \SayCan \cite{MichaelDosaycan} to ours on the \SayCan set of instructions. The best values for a given instruction family are in bold.}
    \label{tab:SayCan}
\end{table}
\end{comment}

\begin{table*}

\newcolumntype{a}{>{\color{OliveGreen}}M{1.3cm}}
\iffinal
    \newcolumntype{a}{M{1.3cm}}
\fi
    \centering
    \begin{tabular}{|M{2.2cm}|M{1.2cm}|a|a|a|a|M{1.2cm}|M{1.4cm}|M{1.2cm}|M{1.4cm}|}\hline
         Instruction Family& \SayCan (plan) & \GPTIII As Planner & \GPTIII As Planner+A & \GPTIV As Planner & \GPTIV As Planner+A&  \GPTIII Ours&  \GPTIII Ours (Auto) & \GPTIV Ours& \GPTIV Ours (Auto)\\ \hline 
         NL Primitive &  0.93&0.47&0.53&0.93&0.93& 0.73&  0.73& \textbf{1.00} & \textbf{1.00}\\ \hline 
         NL Verb &  0.60&0.00&0.00&0.67&0.87&0.27&  0.33& 0.93 & \textbf{1.00}\\ \hline 
         NL Noun &  0.93&0.13&0.07&0.26&0.20&0.27& 0.40 & 0.93 & \textbf{1.00}\\ \hline 
         Structured&  \textbf{0.93}&0.20&0.13&0.87&0.60&0.00&  0.13& 0.20 & \textbf{0.93}\\ \hline 
         Embodiment&  0.64&0.09&0.00&0.55&0.55&0.64&  0.64& 0.82 & \textbf{1.00}\\ \hline 
         Crowd-Sourced&  0.73&0.13&0.07&\textbf{0.93}&0.73&0.27& 0.33 & 0.73 & \textbf{0.93}\\ \hline 
         Long-Horizon&  0.73&0.00&0.00&0.40&0.33&0.20&  0.33 & 0.80 & \textbf{1.00}\\ \hline 
         Drawer&  \textbf{1.00}&0.00&0.00&0.00&0.00&0.66&  0.33& \textbf{1.00} & \textbf{1.00}\\ \hline
         \rowcolor{lightgray} Average& 0.81&0.14&0.12&0.66&0.59&0.34& 0.40& 0.78&\textbf{0.98}\\ \hline
    \end{tabular}%
    \caption{Ablation results of the planning success rate with our \emph{Plan Tool} with different versions
of GPT and with(Auto) or without automatic self-corrections on the \SayCan instruction set. \added{\emph{GPT-X as Planner} refers to the naive baseline of using the LLM directly as the planner, \emph{GPT-X as Planner+A} refers to the same planner with additional context information about affordances added to the prompt.}}
    \label{tab:SayCanOracle}
\end{table*}

\added{As an additional baseline, we compare our approach to a naive \LLMasPlanner implementation that generates a plan based on a textual representation of the initial scene state, a description of the available actions, and the user-specified task in natural language. To assess whether the affordance information alone improves the planning capabilities of the LLM, we compare it against a second \LLMasPlanner implementation that additionally has information about the objects' affordances in the prompt.}

\added{As can be seen in \autoref{tab:SayCanOracle}, our method outperforms the naive \LLMasPlanner implementation for both \GPTIII and \GPTIV, confirming the findings of \cite{liu2023llmp} that the \LLMwPlanner paradigm is superior. Contrary to our expectations, the addition of affordance information decreased the performance of the \LLMasPlanner version. This could be explained by the LLM being overwhelmed by the additional affordance information, which was responsible for almost half of the prompt length, and thus lost track of the task at hand. This shows that directly providing affordance information to the LLM is not sufficient, but using the affordance information in a rule-based system like a classical planner -- as we do -- is a more effective approach.}

Moreover, our method performs equally or better than \SayCan in all categories when using \GPTIV but performs worse when using \GPTIII. Utilizing \GPTIV, our approach outperforms \SayCan, especially in the \emph{Embodiment} and \emph{Long-Horizon} instruction categories. \added{The \emph{Embodiment} category refers to scenarios where the robot's current state needs to be considered, whereas \emph{Long-Horizon} refers to tasks that need more than 7 actions to fulfill.} This is likely owed to our method solely creating a goal state from the user's statement rather than the entire plan. Thus, instructions that require knowledge of the robot's position or current state, or those that require extensive planning, are not as constrained by LLM's limited reasoning capabilities, as the planner generates the plan in a rule-based manner. It should be clarified that \SayCan is designed to optimize the plan's execution, not just the plan itself. Furthermore, we utilize a more recent LLM as opposed to \SayCan's use of PaLM. \added{Also, as can be seen by the leap in success rate from \GPTIII to \GPTIV, the used LLM has a significant influence on the performance of our method. Therefore, it is not possible to directly compare to the approach of \SayCan and make absolute statements about the superiority of one method over the other as long as different LLMs are used as a backbone for planning. However, as PaLM is not publicly available, we are not able to provide an evaluation that makes a direct comparison possible. }

We can see from \autoref{tab:plan} that the planner performs without failure for the \emph{Simple Task} and \emph{Simple Goal} subsets but has more problems with complex goals and the more vague tasks in the \emph{Knowledge} and \emph{Implicit} subsets. \GPTIII seems to have even more problems interpreting vague user instructions to translate them to goals. Additionally, as the planner always finds the minimal plan for the generated goal, most of the found plans are also minimal. The reason for a generated plan not being minimal is primarily a generated goal that is too restrictive. For example, if the task is "Bring me an apple or a banana" and the generated goal is {\footnotesize \texttt {inhand apple0 human0}} instead of \mbox{\footnotesize \texttt{or (inhand apple0 human0) (inhand banana0 human0)}}, the generated plan will not be minimal if it requires more actions to bring the banana than the apple. \added{The average planning time of our approach in the \SayCan set of instructions was 3.3 seconds with \GPTIII and 15,4 seconds with \GPTIV. The scene complexity had a significant influence on the planning time of the \emph{Fast Downward} planner and only a minor influence on the inference time of the LLM. Comparison between the \emph{Simple Goal} and \emph{Complex Scene} sets show that increasing the object number from 30 to 100 only marginally increases the average LLM inference time from 2.7 to 2.8 seconds for \GPTIII and from 19.1 to 21.8 seconds for \GPTIV. The overall planning time increases from 8.4 to 31.6 for \GPTIII and from 28.0 to 59.4 seconds for \GPTIV. This shows that the bottleneck for the planning time in more complex scenes is the planner and not the LLM.}

The results also show the efficiency of self-correction with external feedback, \added{which is a central improvement of our work over \LLMP. The \emph{Plan Tool} without self-correction can be seen as an implementation of \LLMP that generates the initial state of the \PDDL problem directly from the environment representation instead of natural language. It} improves the success rate slightly from 0.32 to 0.49 for \GPTIII and 0.79 to 0.81 for \GPTIV. \replaced{Notably, \GPTIV is so good at translating simple tasks and goals that it does not even need self-correction, as it only performs worse for one scenario in the two sets. This shows that the current auto-correction is only helpful with \GPTIII, and with \GPTIV, the difference is insignificant in our dataset. This could be improved by more specific error messages with more direct hints to improve the previous answer.}{The improvement is rather insignificant on our dataset,} however, on the \SayCan instruction set, the difference between \GPTIV with and without self-correction is far more significant, showing an improvement from 0.78 to 0.98, as can be seen in \autoref{tab:SayCanOracle}. It is mostly caused by the \emph{Structured Language} instruction family and can be mostly explained by semantic error correction. For example, for the task "Pick up the apple and move it to the trash", \ChatGPT answers with \\
\centerline{\footnotesize \texttt{and (inhand apple0 robot0) (in apple0 trash\_can0)}} \\
which is recognized by the semantic error detection and responded with the error message "There is a logical contradiction in the goal. An object that is in the hand of an agent cannot be in another hand or at another place. Please correct your answer". \ChatGPT then corrects its answer to\\
\centerline{\footnotesize \texttt{in apple0 trash\_can0}.}

Overall, the results show that even though simple and explicit tasks can be translated well by \GPTIV, it struggles to correctly interpret the user's intentions when the goal is more indirectly stated. Contextual cues from the environment must be considered to understand the user's goal, as most humans would be able to do. \added{In addition, this also shows that the system as a whole should seek clarification if the user's intentions are unclear.}

\begin{table}
    \centering
    \begin{tabular}{|M{1.15cm}|M{0.5cm} M{0.45cm}|M{0.5cm} M{0.5cm}|M{0.5cm} M{0.45cm}|M{0.5cm} M{0.5cm}|} \hline 
         \multirow{2}{*}{Subset}  &  \multicolumn{2}{c|}{\GPTIII} &  \multicolumn{2}{c|}{\GPTIII Auto} &  \multicolumn{2}{c|}{\GPTIV} & \multicolumn{2}{c|}{\GPTIV Auto} \\ 
         & success & min & success & min & success & min & success & min\\
         \hline 
         Simple Task& 0.70 & 0.63 &  0.70 & 0.63& 0.97& 0.97& \textbf{1.00} & \textbf{1.00}\\ \hline 
         Simple Goal& 0.63 & 0.60&  0.90 & 0.83& \textbf{1.00}& \textbf{0.97}& \textbf{1.00} & 0.93\\ \hline
         \added{ Complex Scene} & \added{0.17} & \added{0.13} & \added{0.77} & \added{0.53} & \added{0.93} & \added{0.87} & \added{\textbf{0.97}} & \added{\textbf{0.93}}\\ \hline
         Complex Goal& 0.23 & 0.17&  0.33 & 0.23& \textbf{0.87} & 0.70& \textbf{0.87} & \textbf{0.73}\\ \hline 
         Knowledge& 0.10 & 0.10&  0.10 & 0.10& 0.53 & 0.53 & \textbf{0.57} & \textbf{0.57}\\ \hline 
         Implicit& 0.10 & 0.10&  0.13 & 0.13& 0.43 & 0.40& \textbf{0.47} & \textbf{0.43}\\ \hline         
         \rowcolor{lightgray} Average& 0.32 & 0.29 &  0.49 & 0.42 & 0.79 & 0.74 & \textbf{0.81} & \textbf{0.77}\\ \hline
    \end{tabular}
    \caption{Ablation results of planning with our \emph{Plan Tool} with different versions of GPT and with or without automatic self-corrections (Auto) on our instruction set. Success refers to the success rate, whereas min refers to the rate of plans that had the minimal length possible for the given goal.}
    \label{tab:plan}
\end{table}

\subsection{AutoGPT+P}

As the planning process for \AutoGPTP involves several steps beyond just planning, using previous scenarios and metrics alone is inadequate. \added{For this reason, we do not only evaluate \AutoGPTP against \SayCan as the scenarios from the \SayCan set of instructions do not include exploration or explicit object substitution.} A crucial aspect of \AutoGPTP is selecting the appropriate tool for each situation and only calling tools other than the \emph{Plan Tool} if they are not in the scene. Therefore, we have incorporated an evaluation metric to determine the optimal number of tools and evaluated the rate of successful plans that use the optimal number of tools. This metric is referred to as \emph{minimal tools} in \autoref{tab:autogptp}.

We designed five scenario sets to assess performance, each containing 30 scenarios. Four of these sets concentrate on individual tools, while the final set requires combining all tools to accomplish complicated tasks. We randomly picked scenarios from the prior segment for the \emph{Plan} subset. Meanwhile, we crafted entirely new scenarios for the \emph{Explore} and \emph{Partial Plan} subsets. For the \emph{Explore} set, hints were partially provided regarding the location of objects, such as "Bring me the cucumber from the fridge". The \emph{Suggest Alternative} and \emph{Combined} sets feature the same scenarios with the exception that for the \emph{Combined} set, only the initial location of the robot is explored. The results can be viewed in \autoref{tab:autogptp}.

\begin{table}
    \centering
    \begin{tabular}{|M{1.3cm}|M{0.7cm} M{0.8cm} M{0.8cm}|M{0.7cm} M{0.8cm} M{0.8cm}|} 
        \hline
        \multirow{3}{*}{Subset} & \multicolumn{3}{m{2.3cm}|}{\GPTIII} & \multicolumn{3}{m{2.3cm}|}{\GPTIV} \\ 
         & success & minimal & minimal tools & success & minimal & minimal tools \\
         \hline
         Plan & 0.53 &  0.50 & 0.30 & \textbf{0.87} & \textbf{0.80} & \textbf{0.80}\\ \hline 
         Partial Plan & 0.37 & 0.20 & \textbf{0.23} & \textbf{0.83} & \textbf{0.67} & 0.13\\ \hline 
         Explore & 0.10 & 0.03 & 0.00 & \textbf{0.77} & \textbf{0.23} & \textbf{0.63} \\ \hline 
         Suggest Alternative & 0.13 & 0.13 & 0.03 & \textbf{0.77} & \textbf{0.53} & \textbf{0.73} \\ \hline 
         Combined & 0.13 & 0.07 & 0.10 & \textbf{0.70} & \textbf{0.53} & \textbf{0.47} \\ \hline
         \rowcolor{lightgray} Average & 0.25 & 0.19 & 0.13 & \textbf{0.79} & \textbf{0.55} & \textbf{0.55} \\ \hline
    \end{tabular}%
    \caption{Evaluation of AutoGPT+P in the metrics success rate, minimal plan length, and minimal tool usage rate comparing \GPTIII to \GPTIV. Best values are written in bold.}
    \label{tab:autogptp}
\end{table}

As can be seen from the \emph{Plan} and \emph{Partial Plan} set, introducing a prior tool selection process does not make the performance worse compared to the scenarios from the planning evaluation. With exploration involved, the success rate gets slightly worse, with the most common mistake being planning before having explored all relevant locations. The \emph{Suggest Alternative} set also has a similar lowered success rate, caused mainly by invalid alternative suggestions. This is expected as the success rate of the medium-sized scenes was 0.83 for the \emph{Suggest Alternative Tool}. The success rate for the \emph{Combined} set is 0.07 lower than the \emph{Suggest Alternative} set, which shows that the addition of exploration does lead to a lower success rate than just using the \emph{Suggest Alternative Tool} alone. Reviewing the data, the most common reason for failure is planning before all necessary objects or replacements are determined.

What can also be seen is that the tool usage is often minimal when only one tool needs to be used, with the exception of the \emph{Partial Plan} set where the \emph{Suggest Alternative Tool} is often called or the tool selection gets stuck in a loop of selecting \emph{Partial Plan}. From the \emph{Explore} scenarios, we can observe that when given a hint of the location of an item, the tool selection never fails to explore the correct location. However, when not given any clues, the tool selection seems to randomly explore locations even if, from the name of the location, it can be inferred that the relevant objects are unlikely there. For example, the system tries to explore the window location to search for vegetables.

In contrast to \GPTIV, \GPTIII mostly fails at the tool selection task. For \emph{Plan} and \emph{Partial Plan} sets, it performs in the same success range as in the \emph{Plan Tool} scenarios. However, the tool selection is not optimal even for those, as seen from the \emph{minimal tools rate}. If it needs to use the \emph{Explore} and \emph{Suggest Alternative} tools, it seemingly chooses the tools randomly and thus has a low success rate.

Overall, the results show the viability of the tool selection process to solve tasks with missing objects and partially unexplored scenes. However, issues like preemptive calling of the \emph{Plan Tool} or inadequate calls of the \emph{Suggest Alternative Tool} remain. Furthermore, as the evaluation set is limited in scope, the results should be approached with caution. 

\subsection{Validation Experiments on \armarVI and \armarDE}
To validate the feasibility of our system, we performed several experiments on the humanoid robots \armarVI \cite{Asfour2019} and \armarDE. As our approach primarily focuses on object affordance detection and planning, we made several assumptions to ease the integration on the robot. 
% listing the restrictions /shortcuts taken in the experiments
We relied on predefined object models for manipulation tasks like grasping, placing, and pouring. Additionally, the locations the robot could navigate to and the environment model are entirely known. However, we dynamically detect the locations of all objects that can be manipulated. Furthermore, to detect liquids inside containers, we assume a predefined liquid is in every liquid container. The object relations are estimated based on the related objects' affordances, and the spatial relations of the object poses are estimated by a fine-tuned MegaPose model \cite{labbé2022megapose}.
For object detection, we used the \emph{yolov5} object detector \cite{glenn_jocher_2022_7347926} that was fine-tuned on a predefined object set \added{from \cite{younes2024kitchen}}. \added{For grasping and placing, we used an affordance-based, memory-centered manipulation framework\cite{pohl2024memorycentered}, and} for pouring, we used an affordance keypoint detection method \cite{Gao2023} to detect the opening of the source container and assume that the target container is symmetric, so moving the object's keypoint above the center of the target object is sufficient. We avoided using real liquids during the experiments to avoid damage to the robot.
% experiment setup
We performed four kinds of tasks with \added{five different} formulations of our user requests \replaced{and objects}{each}. These included \pickplace, \handover, \pour, and  \wipe tasks. All of these require different levels of human-robot collaboration. Whereas the robot needs no help executing the pick and place and wiping tasks, it needs to ask for help for the pouring tasks to open the liquid container. In handover tasks, the human and the robot are equally part of the task.
% results
As shown in the supplementary material, our proposed system, \AutoGPTP, exhibits proficiency in generating executable plans on our robot. \added{Out of the 20 real-life scenarios, 15 were planned successfully.}
Subsequent investigation of failure cases revealed that most of these cases could be attributed to false positive detection of objects or the robot's inability to accurately grasp the target object. \added{This highlights the limited resilience of our approach to failures, which needs to be addressed in future work.}
A more in-depth evaluation of \replaced{these cases}{the experiments} can be found in the supplementary material file. 

\addedsubsection{Discussion}
\added{The results of our evaluation show the potential of \AutoGPTP in real-world scenarios. We show that our system can handle difficult planning situations, such as missing objects. Furthermore, our system achieves a high success rate in transforming user-specified tasks into valid plans. For explicitly stated user goals, it has a planning success rate of nearly 100\%. However, when the user-specified task is more vague, it is mostly limited by its inability to assess the uncertainty in the user's intent and ask for clarification. Another limitation of \AutoGPTP's application in the real world is that everything is modeled in a deterministic way, so uncertainties caused by object recognition, the LLM-generated \OAM, or the unclear user-specified tasks are not taken into account. Furthermore, we do not incorporate any feedback (\eg about skill failures) from execution into our approach, which makes it prone to errors when executed in real-world scenarios. In our evaluation, we also did not consider plans longer than 20 steps, but \cite{liu2024delta} provides evidence that \LLMP, and thus our approach, cannot effectively solve problems that require longer plans.}

\section{Conclusion and Future Work}

In this work, we propose representing objects in the scene as a set of object-affordance-pairs. The scene representation is generated through combined object detection and \OAM (OAM), where object classes are associated with their affordances. Our work demonstrates the utility of \ChatGPT in automatically deriving an OAM for novel classes based on a fixed set of predefined affordances. On our newly proposed set of affordances for planning, we achieved an F1-score of 89\%.

We utilized the scene representation in \AutoGPTP, our proposed planning system, which uses the concept of affordances for planning and alternative suggestions. It consists of an LLM-based tool selection loop that chooses from one of four tools to solve the user-specified task: \emph{Plan}, \emph{Partial Plan}, \emph{Explore}, and \emph{Suggest Alternative}. The \emph{Suggest Alternative Tool} uses the affordances of a missing object to steer the LLM during the alternative suggestion process. Additionally, the \emph{Plan} and \emph{Partial Plan Tool} utilize the LLM to produce goal states in an affordance-based planning domain and generate a plan fulfilling the (partial) goal with a classical planner. The experiments demonstrate that the \emph{Plan Tool} \replaced{vastly surpasses \SayCan on \SayCan's set of instructions in terms of planning when not considering execution, improving the success rate from 81\% to 98\%. The self-correction of semantic and syntactic errors has a significant influence on this, raising the success rate from 79\% to 98\% when compared to the method without self-correction.}{vastly outperforms the naive baseline of a naive \LLMasPlanner implementation. The self-correction of semantic and syntactic errors has a significant influence, raising the success rate from 78\% to 98\% when compared to the method without self-correction.}

Furthermore, our affordance-guided \emph{Suggest Alternative Tool} outperforms a naive approach in scenes with 20 and 70 objects by 13\%. When evaluating the system's overall performance, we reach an average success rate of 79\% on our dataset containing 150 tasks. Difficulties persist mainly due to the LLM selecting incorrect tools. Therefore, a reevaluation of the tool selection process is necessary to address this issue.

Our validation experiments show that the generated plans can be successfully executed on the robot and that the symbolic representation of objects from the planning domain can be transferred to the subsymbolic object representations needed for skill execution\added{, however, our experiments also showed the low resilience to failures during execution.}

In future work, probabilistic aspects should be integrated for improved accuracy in real-world deployment. This involves representing the OAM as a probabilistic function that can be updated incrementally based on user feedback or execution. \added{Alternatively, direct verbal corrections from the human like "You cannot use a fork for cutting." can also be taken into account to update the probability of a given affordance. The system should also decide whether to retry a failed action or generate a different plan instead based on the updated confidence of the affordance. This could reduce the error rate during execution.} Additionally, a probabilistic representation of the object-affordance-pairs, which includes the confidence level from the object detection, can be combined with this. The resulting probabilistic scene representation can then be used in conjunction with a planner that optimizes the probability of a plan's success rather than just plan length.

Furthermore, a more versatile human-robot interaction would be beneficial. This involves equipping the system with the capability to seek clarification if the user's instruction is unclear and granting the user the ability to modify or terminate the plan during execution.

\section*{Acknowledgments}
\ackHariaEuRobinJuBot

\bibliographystyle{plainnat}
\bibliography{bibliography}
\balance 
\clearpage

\onecolumn
\appendix

\subsection{Proposed Affordances for Planning}

The following table lists all affordances that were detected using our automatic OAM using ChatGPT. Affordances in bold are used in the planning domain that is used for evaluation.
\begin{multicols*}{2} 

\begin{tabular}{|
    p{\dimexpr.25\columnwidth-2\tabcolsep}|
    p{\dimexpr.7\columnwidth-2\tabcolsep}|
}
\hline
Affordance & Description \\
\hline
\hline
\textbf{grasp} & The object can be grasped in any way\\
\hline
\textbf{carry} & The object can be carried with two hands\\
\hline
assisted-carry & Two or more people can carry the object cooperatively without being in each other's way\\
\hline
\textbf{cut} & The object can be used to cut other objects\\
\hline
\textbf{contain} & The object is designed to put either objects or liquids inside of it\\
\hline
\textbf{liquid-contain} & The object is designed to put liquids inside of it\\
\hline
\textbf{enclosed-contain} & The object can be closed so the objects stored inside it do not fall or leak out when moving\\
\hline
\textbf{pour} & The object can be used to pour liquids\\
\hline
\textbf{precise-pour} & The object can be used to precisely pour liquids into small containers like glasses\\
\hline
\textbf{drink} & The object is designed to drink from \\
\hline
constrained-move & The object can only be moved with restrictions as it is mounted to another object like a door\\
\hline
rotate & The object can be turned\\
\hline
axis-roll & The object can be rolled around an axis\\
\hline
free-roll & The object can be rolled freely in any direction as it is approximately sphere-shaped \\ 
\hline 
push & The object can be pushed away from oneself\\
\hline
pull & The object can be pulled towards oneself\\
\hline
\textbf{open} & The object can be opened\\
\hline
\textbf{close} & The object can be be closed\\
\hline
\textbf{support} & The object provides good support for other objects standing on it\\
\hline
stack & The object can be stacked on objects of the same type \\ 
\hline
\end{tabular}
\\[3cm]
\begin{tabular}{|
    p{\dimexpr.25\columnwidth-2\tabcolsep}|
    p{\dimexpr.7\columnwidth-2\tabcolsep}|
}
\hline
Affordance & Description \\
\hline
\hline
\textbf{sturdy-support} & The object supports other objects on it and it allows to cut objects on top of it\\
\hline
vertical-support & The object can be leaned against safely\\
\hline
scoop & The object can be used to scoop or shove material like powder or objects\\
\hline
stir & The object can be used as a tool to stir\\
\hline
distance-connect & The object physically connects other objects without the connected objects needing to be in contact\\
\hline
contact-connect & The object connects objects together\\
\hline
pierce & The object can be used to pierce through other objects\\
\hline
pick & The object can be used to pierce other objects to pick them up\\
\hline
hit & The object can be used to hit other objects with\\
\hline
pound & The object can be swung to pound other objects\\
\hline
swing & The object can be swung to hit other objects with\\
\hline
dry-swipe & The object can be used to wipe dust or rubble efficiently\\
\hline
\textbf{wet-swipe} & The object can be used to swipe other objects with water\\
\hline
\textbf{heat} & The object can be used to make other objects or liquids warmer\\
\hline
\textbf{heat-resistance} & The object can be safely exposed to temperatures over 100 degrees celcius\\
\hline
\textbf{liquid} & The object is a liquid\\
\hline
\textbf{drinkable} & The object can be safely drunk by a human\\
\hline
\textbf{consumable} & The object can be safely consumed by a human\\
\hline
\end{tabular}
\end{multicols*}

\subsection{Description of Experiments on \armarVI and \armarDE}

The aim of the experiments is to validate that generated plans can be executed on a real robot. The actually executed actions are subject to improvement but work as a proof of concept for mapping from symbolic to subsymbolic actions.
Failure cases during the experiments were mostly caused by object detection and object localization. To make the execution process more straightforward we ignored classes that were often detected as false positives. As the main focus of this work is not object detection, this should be justifiable. We plan to integrate a more robust object detector and localizer into our approach to be less prone to errors.
When there were no false negatives or positives our method reliably generated valid plans with only a few failure cases. For example, the goal state for "Bring me a coffee cup" was generated to be and \texttt{(in-hand coffee\_cup0 robot0) (at robot0 human0)}. The results are described in the following notation:\\
\texttt{Task:} Describes the user-specified task in natural language (With other formulations that resulted in the same plan, but are not shown in the video in brackets)\\
\texttt{Locations:} Describes known locations that can be explored\\
\texttt{Initial State:} Describes the initial state unknown to the robot (except for agent locations). Object relations are discovered by object detection.\\
\texttt{Generated Plan:} Used Tools with parameters, for the PLAN Tool the generated goal and plan are listed below, for SUGGEST ALTERNATIVE the suggested alternative to the missing object is listed below in the notation "missing \verb|->| alternative"\\
\texttt{Failures:} user-specified tasks that should result in the same plan but do not (due to reasons stated in brackets). We only describe failure cases due to detection and planning as failure cases due to execution are independent of our approach. For execution failures we simply restarted the experiment.\\

\begin{enumerate}
\item
\textbf{Pick and Place:}
\begin{quote}
	\texttt{\textbf{Task:}} "Put the sponge next to the screwbox" ("Put the sponge on table1", "Put the sponge on the other table")\\
	\texttt{\textbf{Locations}: {\footnotesize table0, table1}}\\
	\texttt{\textbf{Initial State}:
{\footnotesize \begin{quote} 
 at robot0 table1, on sponge0 table0, on tea\_packaging0 table0, on tea\_packaging1 table0, on milk\_box0 table0, on coffee\_cup0 table0, liquid\_in milk0 milk\_box0, closed milk\_box0, on screw\_box0 table1, on spraybottle0 table1, on grease0 table1, on soap0 table1
 \end{quote}}}
	\texttt{\textbf{Generated Plan}:
    {\footnotesize \begin{quote} 
        EXPLORE table0\\
		PLAN (for goal: on sponge0 table1)
        \begin{quote}
          grasp robot0 sponge0 table0 left \textnormal{(Robot grasps sponge from table0 with left hand)}\\
		   move robot0 table0 table1 \textnormal{(Robot moves from table0 to table1)}\\
		   place robot0 sponge0 table1 left \textnormal{(Robot places sponge on table1 with the left hand)}
        \end{quote}
     \end{quote}}}
    \texttt{\textbf{Failures}}:
    \begin{quote}
        "Put the sponge next to the spraybottle" (due to misdetected spraybottle)\\
        "Put the sponge on the table in front of you" (the PLAN tool does not know the initial position before exploration, so the wrong table is set as the target)
    \end{quote}
\end{quote}

\item  
\textbf{Handover:}
\begin{quote}
	\texttt{\textbf{Task:}} "Give me a glass" ("Fetch me a glass, "I want to have a glass", "Hand me a glass over")\\
	\texttt{\textbf{Locations:} {\footnotesize table0, human0}}\\
	\texttt{\textbf{Initial State:}
 {\footnotesize \begin{quote} at robot0 human0, on coffee\_cup0 table0, on milk\_box0 table0, liquid\_in milk0 milk\_box0, closed milk\_box0
  \end{quote}}}
	\texttt{\textbf{Generated Plan:}
 {\footnotesize \begin{quote} 
		EXPLORE table0\\
		SUGGEST ALTERNATIVE glass
  \begin{quote}
          glass -> coffee\_cup
          \end{quote}
		PLAN (for goal: in-hand coffee\_cup0 human0)
  \begin{quote}
		   grasp robot0 coffee\_cup0 table0 left \textnormal{(Robot grasps coffee cup from table0 with left hand)}\\
		   move robot0 table0 human01 \textnormal{(Robot moves from table0 to the human)}\\
		   handover robot0 human0 coffee\_cup left \textnormal{(Robot gives human the coffee cup)}
     \end{quote}
 \end{quote}}}
    \texttt{\textbf{Failures:}}
     \begin{quote}
      "Bring me a glass" (Results in goal state {\footnotesize \texttt{and (in-hand coffee\_cup0 robot0) (at robot0 human0)}})
      \end{quote}
\end{quote}

\newpage
      
\item
\textbf{Pouring:}
\begin{quote}
	\texttt{\textbf{Task:}} "I want a glass of water" ("Pour me a glass of water", "I want to drink water", "I am thirsty")\\
	\texttt{\textbf{Locations:} {\footnotesize table0}}\\
	\texttt{\textbf{Initial State:} 
 {\footnotesize \begin{quote}
  at robot0 table0, at human0 table0, on coffee\_cup0 table0, on milk\_box0 table0, liquid\_in milk0 milk\_box0, closed milk\_box0
  \end{quote}}}
	\texttt{\textbf{Generated Plan:}
 {\footnotesize 
     \begin{quote} 
		SUGGEST ALTERNATIVE glass
          \begin{quote}
          glass -> coffee\_cup
          \end{quote}
		SUGGEST ALTERNATIVE water
          \begin{quote}
          water -> milk
          \end{quote}
		PLAN (for goal: liquid\_in milk0 coffee\_cup0)
          \begin{quote}
		    open human0 milk\_box0 left \textnormal{(Robot asks human to open the milk box for it)}\\
		    grasp robot0 milk\_box0 table0 right \textnormal{(Robot grasps milk box from table0 with right hand)}\\
		    pour robot0 milk\_box0 milk0 coffee\_cup0 right \textnormal{(Robot pours milk from the milk box to the coffee cup)}
         \end{quote}
   \end{quote}}}
    \texttt{\textbf{Failures:}}
     \begin{quote}
        "I want a cup of water" (LLM keeps calling SUGGEST ALTERNATIVE coffee\_cup0, this is an entirely wrong usage of this tool)
\end{quote}
\end{quote}
        
\item		  
\textbf{Wiping:}
\begin{quote}
	\texttt{\textbf{Task:}} "I spilled milk on this table" ("Clean this table", "Wipe this table", "This table is dirty")\\
	\texttt{\textbf{Locations}: {\footnotesize table0, table1}}\\
	\texttt{\textbf{Initial State:} 
 {\footnotesize \begin{quote} 
 at robot0 human0, at human0 table0, on screw\_box0 table1, on spraybottle0 table1, on grease0 table1, on soap0 table1, on sponge0 table1
  \end{quote}}}
	\texttt{\textbf{Generated Plan:}
 {\footnotesize \begin{quote} 
		EXPLORE table1\\
		PLAN (for goal: clean table0)
        \begin{quote}
		  grasp robot0 sponge0 table1 left \textnormal{(Robot grasps sponge from table1 with left hand)}\\
		  move robot0 table1 table0 \textnormal{(Robot moves from table1 to table0)}\\
		  wipe robot0 table0 sponge left \textnormal{(Robot wipes table0 with sponge in left hand)}
    \end{quote}
 \end{quote}}}
    \texttt{\textbf{Failures:}}
     \begin{quote}
       "I spilled my milk on the table" (LLM generates goal state {\footnotesize \texttt{clean table1}}, which is clearly not meant as the human is at table0.)
       \end{quote}
\end{quote}

\end{enumerate}

\end{document}